\documentclass[a4paper,fleqn]{cas-sc}
\usepackage{amssymb}
\usepackage[numbers]{natbib}
\usepackage{hhline}
\usepackage{threeparttable}
\usepackage{tabularx}
\usepackage{multirow}
\usepackage{xcolor}
\usepackage{booktabs}
\usepackage{multicol}
\usepackage[ruled,vlined]{algorithm2e}
\usepackage[figuresright]{rotating} 
\usepackage{subfigure}
\usepackage{graphicx}
\usepackage{colortbl}
\usepackage{color}
\usepackage[numbered]{bookmark}
\usepackage{hyperref}

\hypersetup{
	bookmarksnumbered=true,
	bookmarksopen=true,
	bookmarksopenlevel=5,
	pdfstartview=Fit,
	pdfpagemode=UseOutlines
}
\usepackage{array}

\usepackage[switch]{lineno}
\def\tsc#1{\csdef{#1}{\textsc{\lowercase{#1}}\xspace}}
\tsc{WGM}
\tsc{QE}
\tsc{EP}
\tsc{PMS}
\tsc{BEC}
\tsc{DE}
\begin{document}
\let\WriteBookmarks\relax
\def\floatpagepagefraction{1}
\def\textpagefraction{.001}
\shorttitle{A Distance Similarity-based Genetic Optimization Algorithm for Satellite Ground Network Planning Considering Feeding Mode}
\shortauthors{Yingying Ren et~al.}

\title [mode = title]{A Distance Similarity-based Genetic Optimization Algorithm for Satellite Ground Network Planning Considering Feeding Mode}                      
%\tnotemark[1,2]

%\tnotetext[1]{This document is the results of the research
	%   project funded by the National Science Foundation.}
%
%\tnotetext[2]{The second title footnote which is a longer text matter
	%   to fill through the whole text width and overflow into
	%   another line in the footnotes area of the first page.}

\author[1,2,3]{Yingying Ren}
\ead{yingyingren@gxu.edu.cn}
\address[1]{School of Computer, Electronics and Information, Guangxi University, Nannig, China}
\address[2]{Guangxi Key Laboratory of Multimedia Communications and Network Technology}
\address[3]{Key Laboratory of Parallel, Distributed and Intelligent Computing (Guangxi University)}

\author[1]{Qiuli Li}
\ead{qiuli_93@st.gxu.edu.cn}

\author[4]{Yangyang Guo}
\ead{g2002dmu@163.com}
\address[4]{School of Systems Science, Beijing Jiaotong University, Beijing, China}

\author[5,6,7]{Witold Pedrycz}
\ead{wpedrycz@ualberta.ca}
\address[5]{Department of Electrical and Computer Engineering, University of Alberta, Edmonton, Canada}
\address[6]{Systems Research Institute, Polish Academy of Sciences, Poland}
\address[7]{Faculty of Engineering and Natural Sciences, Department of Computer Engineering, Sariyer/Istanbul, Turkiye}

\author[8]{Lining Xing}
\ead{lnxing@xidian.edu.cn}
\address[8]{School of Electronic Engineering, Xidian University, Xian, China}

\author[9]{Anfeng Liu}
\ead{afengliu@mail.csu.edu.cn}
\address[9]{School of Electronic Engineering, Central South University, Changsha, China}

\author[10]{Yanjie Song}
\ead{songyj_2017@tju.edu.cn}
\cormark[1]
\address[10]{National Engineering Research Center of Maritime Navigation System, Dalian Maritime University, Dalian, China}
\cortext[cor1]{Corresponding author}

%\cormark[1]
%\ead{cvr_1@tug.org.in}
%
%
%\credit{Conceptualization of this study, Methodology, Software}
%
%\address[1]{Elsevier B.V., Radarweg 29, 1043 NX Amsterdam, The Netherlands}
%
%\author[2,4]{Han Theh Thanh}[style=chinese]
%
%\author[2,3]{CV Rajagopal}[%
%   role=Co-ordinator,
%   suffix=Jr,
%\ead{cvr3@sayahna.org}
%\ead[URL]{www.sayahna.org}
%
%\credit{Data curation, Writing - Original draft preparation}
%
%\address[2]{Sayahna Foundation, Jagathy, Trivandrum 695014, India}
%
%\author%
%[1,3]
%{Rishi T.}
%\cormark[2]
%\fnmark[1,3]
%\ead{rishi@stmdocs.in}
%\ead[URL]{www.stmdocs.in}
%
%\address[3]{STM Document Engineering Pvt Ltd., Mepukada,
	%    Malayinkil, Trivandrum 695571, India}

%\fntext[fn1]{This is the first author footnote. but is common to the third
	%  author as well.}
%\fntext[fn2]{Another author footnote, this is a very long footnote and
	% It should be a long footnote. But this footnote is not yet
	%  sufficiently long enough to make two lines of footnote text.}
%
%\nonumnote{This note has no numbers. In this work, we demonstrate $a_b$
	%  the formation Y\_1 of a new type of polariton on the interface
	%  between a cuprous oxide slab and a polystyrene micro-sphere placed
	%  on the slab.  }

\begin{abstract}
With the rapid development of the satellite industry, the information transmission network based on communication satellites has gradually become a major and important part of the future satellite ground integration network. However, the low transmission efficiency of the satellite data relay back mission has become a problem that is currently constraining the construction of the system and needs to be solved urgently. Effectively planning the task of satellite ground networking by reasonably scheduling resources is crucial for the efficient transmission of task data. In this paper, we hope to provide a task execution scheme that maximizes the profit of the networking task for satellite ground network planning considering feeding mode (SGNPFM). To solve the SGNPFM problem, a mixed-integer planning model with the objective of maximizing the gain of the link-building task is constructed, which considers various constraints of the satellite in the feed-switching mode. Based on the problem characteristics, we propose a distance similarity-based genetic optimization algorithm (DSGA), which considers the state characteristics between the tasks and introduces a weighted Euclidean distance method to determine the similarity between the tasks. To obtain more high-quality solutions, different similarity evaluation methods are designed to assist the algorithm in intelligently screening individuals. The DSGA also uses an adaptive crossover strategy based on similarity mechanism, which guides the algorithm to achieve efficient population search. In addition, a task scheduling algorithm considering the feed-switching mode is designed for decoding the algorithm to generate a high-quality scheme. The results of simulation experiments show that the DSGA can effectively solve the SGNPFM problem. Compared to other algorithms, the proposed algorithm not only obtains higher quality planning schemes but also has faster algorithm convergence speed. The proposed algorithm improves data transmission profits for satellite ground networking by achieving higher and more stable performance. Moreover, this similarity evaluation approach is applicable to solving other types of communication satellite scheduling problems.
	
\end{abstract}

%\includegraphics{figs/grabs.pdf}

%\item research highlights item 1
%\item research highlights item 2
%\item research highlights item 3


\maketitle

\section{Introduction}

In recent years, with the continuous development of aerospace technology, satellites have been required to communicate with remote ground stations in the fields of information communication, environmental detection, and space navigation. Satellite communication technology has become an important part of modern communication systems. Communications satellites have built a comprehensive communications satellite network through the coverage of multiple satellites, the construction of ground station networks, and other types of resources. The advancement of communication satellite technology not only addresses intricate challenges across various domains, but also witnesses a growing demand for diverse communication applications in associated industries, thereby facilitating the provision of enhanced and convenient services. In different fields, the successful application of communication satellites is intricately linked to a pivotal technology known as data return. Data return technology is crucial in satellite communications, as it transmits data collected by satellites back to ground stations, ensuring high-quality and stable communication services. Due to communications satellites operating in different orbits, data return encounters various challenges during the execution of satellite missions. For example, in low Earth orbits, the high velocity of communication satellites results in reduced visibility times relative to the ground and necessitates frequent switching of communication links, thereby imposing greater demands on data return timeliness. Regarding medium orbit, medium Earth orbit satellites exhibit a broader ground coverage in comparison to low Earth orbit satellites; however, they are more susceptible to signal transmission delays and attenuation issues. Therefore, how to obtain a reasonable return scheme is particularly important for the system.

    The communication satellite system can  effectively allocate and assign communication satellites to fulfill diverse communication tasks concurrently, in accordance with the specific requirements of each task. To meet the mission requirements, the communication satellite system must carefully consider the establishment and stability of inter-satellite and satellite-to-ground communication links. The reliability and stability of these links directly determine the efficacy of data transmission between the satellite and ground station, which is pivotal for successful data uploading or downloading. Therefore, the communication satellite system must ensure that communication links between satellites and between satellites and the ground can be effectively established and maintained to support the smooth execution of various communication tasks. In practical scenarios, satellites can only fly around the earth in fixed orbits, and the location of the ground station is also fixed, which makes the construction time of the communication link strictly limited. This means that the total length of time that can be used to build communication links during the day is limited, and these time segments need to be fully utilized to ensure the timeliness and stability of the data transfer task. The introduction of feed-switching mode of operation can effectively solve the problems of inefficient data transmission and unstable signal communication connection. By being equipped with a dedicated feeding antenna, the satellite can transmit additional energy to the satellite, allowing a new task to be executed immediately after one. According to different demands and conditions, the satellite can intelligently execute link switching and dynamically select the optimal antenna for communication, which enables the communication system to adapt to changing communication demands and network conditions, execute communication tasks more efficiently, and provide more reliable data transmission services. However, even with the introduction of the new feed-switching mode, there will still be tasks that cannot be performed due to various limitations and constraints. How to extend the total time of the constructed communication link to make the high-quality task successfully executed in this case has become the optimization direction in the satellite ground network planning considering the feeding mode. \textcolor[rgb]{0,0,0}{In a word, satellite ground network planning considering feeding mode (SGNPFM) is to develop a rational satellite network construction program based on the capabilities of satellites and ground station resources.} 

Currently, the SGNPFM problem lacks substantial research, and the existing studies are neither sufficiently mature nor comprehensive. As a classical problem in the field of satellite mission planning, satellite range scheduling problem (SRSP) has been widely studied by scholars all over the world in recent years. These two kinds of problems involve antenna acquisition and link construction between satellite and ground station, which can bring some inspiration to our research. Ou et al. proposed a MIP model with task sequence profit maximization as the optimization objective \cite{ou2023101233}. Song et al. constructed a Bi-objective optimization model considering the objectives of task profit and task failure rate \cite{song2019learning}. Du et al. gave a three-objective optimization model with the objective of minimizing the degree of request failure and resource load balancing for a given schedule \cite{du2019moea}. The mathematical planning model for SGNPFM will be further developed based on the existing SRSP model, incorporating considerations for link establishment requirements and the operational characteristics of communication satellites. In SGNPFM, when a communication satellite exceeds the acceptance range of a conventional ground station antenna, the feeding mode of operation enables swift establishment of a communication link with a new ground station antenna through adjustment of antenna orientation. This approach facilitates more efficient provision of communication services and extends mission execution duration \cite{chen2024data}. However, this method poses significant algorithmic design challenges as it necessitates the ability to select from numerous task combination relationships for constructing sustainable link-building schemes over prolonged periods.

Satellite ground network planning considering feeding mode (SGNPFM) is similar to satellite range scheduling problem (SRSP) \cite{du2019moea}, job scheduling problem (JSP), vehicle routing problem (VRP) \cite{li2003local}, and other classical combinatorial optimization problems. The in-depth exploration of such issues is crucial to the significant improvement of the working efficiency of communication satellites. Dai et al. considered the feature points and indicators of multi-objective optimization and proposed a design scheme of regional satellite constellation to achieve the coverage problem of communication satellite optimized regional terrestrial-satellite network. Experiments showed that the designed scheme can achieve good coverage of the target areas \cite{dai2018satellite}. Al-Hraishawi et al. proposed a novel load balancing scheduling algorithm considering the impact of carrier aggregation on the scheduling of communication satellite systems \cite{al2021scheduling}. Liu et al. proposed a data-driven parallel adaptive large neighborhood search algorithm for the routing problem of communication satellite constellation, and combined heuristic initialization rules with an improved strategy of adaptive mechanism. The proposed algorithm effectively improved the performance of the average transmission time delay \cite{liu2022data}. To enhance the system sum rate and energy efficiency of satellite communication, Peng et al. proposed a feasible optimization method based on the minimum mean square error criterion and logarithmic linearization for optimizing power allocation among user terminals \cite{peng2021hybrid}. Yin et al. considered the load balancing problem between satellites and different base stations (BSs) in the satellite-terrestrial integrated networks and proposed the priority sampling-based DDQN (PSDDQN) algorithm. This algorithm had been proven to have better utility \cite{yin2022deep}. Wang et al. proposed a dynamic spatio-temporal approximation (DSTA) model and a novel beam collaboration scheduling algorithm, which effectively reduced the interaction overhead between users and satellites in satellite aviation communication \cite{wang2023satellite}. Liu et al. proposed an integrated energy efficiency planning algorithm and an efficient local search strategy \cite{liu2024joint}. Zhang et al. studied multi-satellite cooperative networks with the goal of expanding communication coverage. A hybrid beamforming method and a heuristic user scheduling scheme were proposed. The experiment showed that the proposed scheme significantly improved the network communication performance of users \cite{zhang2024multi}. Liu et al.proposed a knowledge-assisted adaptive large neighborhood search algorithm, which combined a data mining method with IP modeling and effectively solved the satellite-ground link scheduling problem \cite{liu2024knowledge}.

Since the problem of satellite range scheduling was put forward, many algorithms have been studied and applied to solve this problem. The accuracy of the search strategy determines whether these algorithms fall into the category of deterministic or random search algorithms \cite{li2023reviews}. Deterministic algorithm searches the solution space globally, exhibiting the characteristic of consistently generating identical outputs for a given specific input, such as branch and bound algorithm \cite{rigo2022branch}, and dynamic programming \cite{liu2005method}. Random search algorithm refers to those algorithms constructed based on intuition or experience, which give an approximate optimal solution to a problem within an acceptable cost \cite{li2023reviews}. Compared with the deterministic algorithm, the random search algorithm can perform better on large-scale problems in various complex satellite range scheduling problems. Among them, a series of traditional and improved evolutionary algorithms such as genetic algorithm \cite{parish1994genetic}, ant colony algorithm \cite{zhang2018ant}, and particle swarm optimization algorithm \cite{huijing2022resource} have been proposed and widely used.

The evolutionary algorithm demonstrates remarkable efficacy in addressing the satellite mission planning problem, thereby providing valuable insights for investigating the SGNPFM. Zhang et al. proposed an algorithm for multi-satellite control resource scheduling problem based on ant colony optimization (MSCRSP-ACO), which used the idea of updating pheromones in two steps to optimize the solution space of the problem, and also employed several heuristic strategies to guide the ant colony optimization process \cite{zhang2014multi}. Khojah et al. proposed an improved particle swarm optimization algorithm for the satellite scheduling problem of multi-objective optimization and obtained high-quality solutions through task prioritization \cite{zhang2014multi}. Song et al proposed a fireworks dynamic algorithm for generating relay satellite scheme. The algorithm performed well in specific scenes \cite{song2021solving}. Du et al. proposed a memetic algorithm framework based on multi-objective evolutionary algorithm (MOEA-MA), and designed a series of operators for the conflict resolution and load-balance to solve the multi-objective satellite range scheduling problem \cite{du2019moea}. To solve the problems of oversubscribed and sequence dependency in SRSP, Song et al. proposed a learning-based artificial bee colony algorithm, which improved the population structure through two different learning strategies \cite{song2021solving}. Liu et al. proposed an improved whale optimization algorithm. A series of detection and search strategies proposed in the algorithm can effectively improve the utilization of communication satellite resources \cite{liu2022communication}. Song et al. proposed a reinforcement learning-based memetic algorithm (RL-MA) for energy-efficient satellite range scheduling problem and simultaneously generated an ensemble local search strategy (ELSS) by combining multiple random and heuristic operators \cite{song2024energy}.

Among various evolutionary algorithms, the genetic algorithm achieves optimization by simulating the process of population reproduction, which has demonstrated remarkable efficacy in solving combinatorial optimization problems. The successful implementation of various genetic algorithms in addressing the communication satellite mission planning problem, which bears resemblance to our current study, has also provided valuable insights. Wei et al. proposed a tight time-indexed formulation of communication data relay satellite scheduling problem and designed a genetic algorithm based on the problem \cite{wei2008genetic}. Dai et al. designed a multi-objective genetic algorithm for regional coverage to achieve the coverage of regional terrestrial-satellite network \cite{dai2018satellite}. Zhang et al. proposed an improved genetic algorithm (IGA) for satellite imaging and data transmission scheduling problem, which adopted a novel idea of encoding and decoding to match the specific request with the corresponding satellite-ground resources, and designed a series of heuristic operators \cite{zhang2022improved}. To solve the energy-efficient survival problem of LEO communication satellites, Jing et al. proposed an energy-efficient routing scheme based on the genetic algorithm and built the satellite aging model. The algorithm effectively improved the lifetime of communication satellite \cite{jing2023energy}. Petelin et al. focused on the different performances of different multi-objective optimization evolutionary algorithms for satellite communication optimal scheduling problem. By using six different evolutionary multi-objective algorithms to solve the ground station scheduling problem, it was analyzed that the decomposition-based MOEA/D had better performance \cite{petelin2023multi}. Song et al. proposed a knowledge-based genetic algorithm, incorporating knowledge-driven initialization and a series of cross mutation strategies, which exhibited remarkable performance in addressing the relay satellite system mission scheduling problem \cite{song2020knowledge}. Despite its ability to provide excellent solutions for the satellite range scheduling problem, the traditional genetic algorithm exhibits slow convergence speed and its blind search approach often leads to low efficiency or repeated search, thereby posing a challenge in addressing large-scale problems. Therefore, many scholars have optimized and improved the genetic algorithm when studying SRSP, and carried out a lot of research on the treatment of population and operator. Bai et al. proposed a multi-dimensional genetic algorithm for spacecraft TT\&C scheduling problem with different orbits. The algorithm adopted a new coding method and designed the corresponding crossover and mutation operators \cite{bai2018multi}. Song et al. used the k-means clustering method combined with the genetic algorithm framework, and designed the crossover and mutation operators based on clustering to solve the SRSP problem \cite{song2023cluster}. Chen et al. proposed a genetic algorithm based on population perturbation and elimination strategy for multi-satellite TT\&C scheduling problem \cite{chen2021population}. Song et al. combined reinforcement learning with genetic algorithm, and used Q-learning to dynamically adjust the algorithm for neighborhood search operation to solve the electromagnetic detection satellite scheduling problem. Intelligent decision greatly improved the development ability of algorithm \cite{song2023rl}.

Based on the aforementioned study, it is evident that evolutionary algorithms exhibit significant advantages in population-based search and demonstrate superior performance in large-scale optimization problems. Consequently, we have opted to employ an evolutionary algorithm for addressing the SGNPFM problem. However, one challenge faced by evolutionary algorithms is their susceptibility to local optima after extensive search iterations. Additionally, during the solution search process, task-specific data can significantly influence the algorithm's performance. Therefore, discussing both the algorithm itself and its compatibility with task attributes separately would not yield targeted profits for the investigated problems.

Considering the relationship between data in the problem can enhance search results significantly during large-scale search iterations. While previous scholars have also explored the role of data features in traditional evolutionary algorithms, these approaches often lack comprehensiveness in extracting task attribute features. To fully account for task attribute characteristics, we propose incorporating task and individual similarities into the genetic algorithm framework to optimize population iteration and search strategies. To our knowledge, existing research rarely integrates task similarity with evolutionary algorithms to address satellite planning and scheduling problems. This approach enhances overall algorithm convergence and achieves desired outcomes.

The main contributions of our research are as follows.

1. A mixed integer programming model describing the SGNPFM problem is constructed. The mixed integer programming model takes into account many factors such as feed-switching, and puts forward a series of assumptions to ensure the normal execution of the ground station and satellite. The objective function is designed to maximize the profit of the link-building task. We also consider various constraints such as the conditions of feed-switching, ground station antenna capabilities, and mission execution window limitations.

2. A distance similarity-based genetic optimization algorithm is proposed to solve the SGNPFM problem. Combining the attribute characteristics of SGNPFM tasks, we calculates the weighted Euclidean distance similarity between tasks and propose an individual similarity calculation strategy and selection method to guide the optimization process of the population. In addition, an adaptive crossover strategy based on similarity mechanism is used to improve the ability of neighborhood search.

3. A new task scheduling algorithm is specifically designed to generate SGNPFM plans. This algorithm incorporates the feed-switching mode into the conventional satellite planning model, and can quickly identify whether the communication task can be scheduled based on time window correlation.

4. The performance of DSGA is validated in several ways through simulation experiments. Compared to the comparison algorithms, DSGA has excellent performance in solution performance, convergence, search speed, and other aspects. The algorithm proposed in this paper not only provides a better solution for SGNPFM but also provides new ideas for solving other combinatorial optimization problems.

The remainder of this paper is organized as follows. The second section gives the text description and mathematical model of SGNPFM. The third section introduces the overall process and some key steps of the distance similarity-based genetic optimization algorithm in detail. The fourth section verifies the effectiveness of DSGA through simulation experiments. The fifth section summarizes the research of the article and discusses the directions for further research in the future.

\section{Model}
\label{Model}
This section gives a detailed description of satellite ground network planning considering feeding mode, introducing the symbols and variables involved in the model, the assumptions, the objective function, and the constraints.

\subsection{Problem Description}
\begin{figure}[htp]
	\centering
	\includegraphics[width=0.5\textwidth]{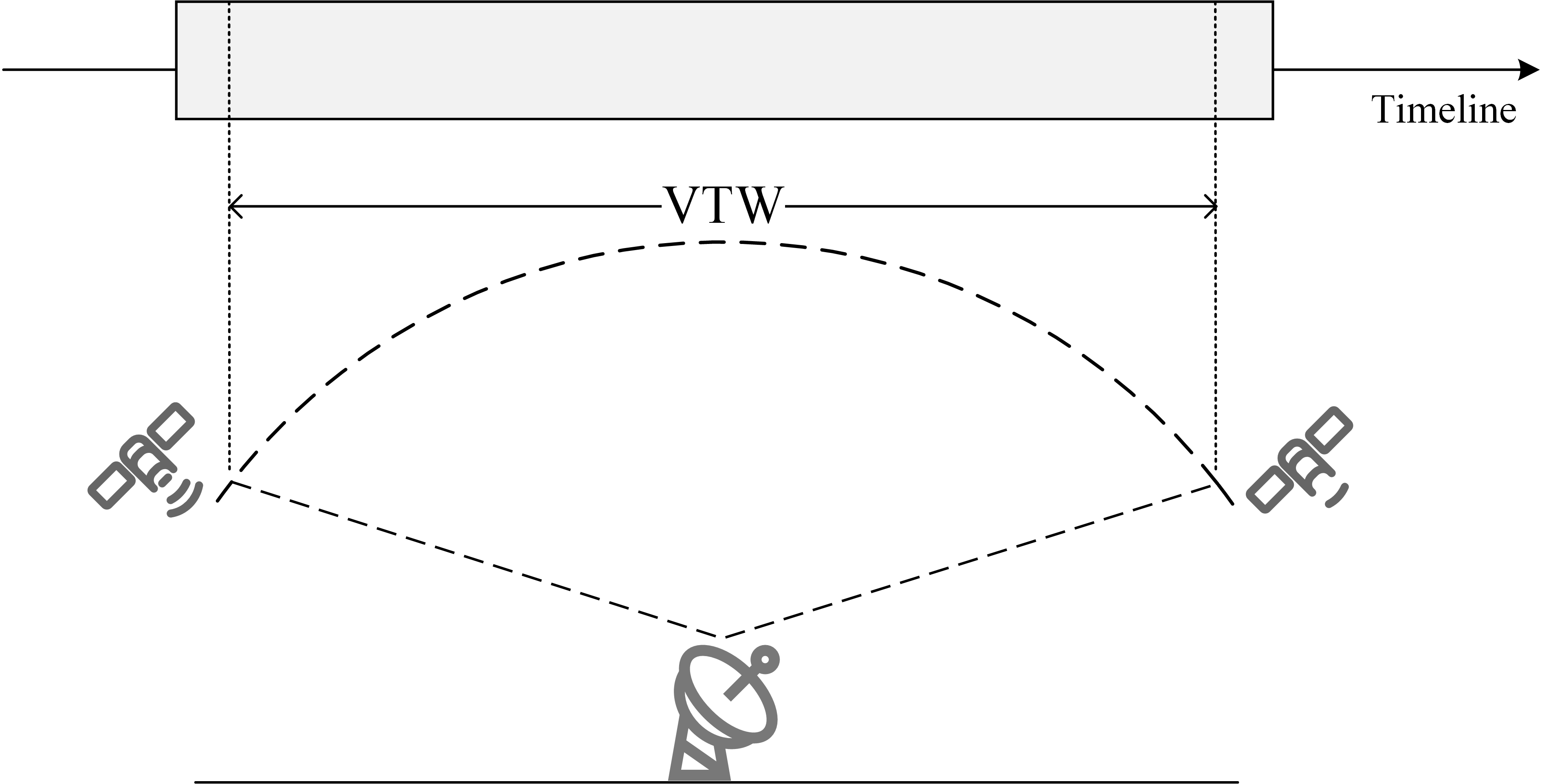}
	\caption{\textcolor[rgb]{0,0,0}{Schematic Diagram of VTW.}}
	\label{Schematic Diagram of VTW}
\end{figure}

In SGNPFM, there are multiple satellite and ground station resources. Satellites orbit the Earth in a fixed orbit, and only when they reach the receiving range of the ground station antenna equipment can they establish a data transmission link to perform the mission. This receiving range is also known as the visible time window (VTW). Fig.\ref{Schematic Diagram of VTW} shows the schematic diagram of a VTW. In the feed-switching mode of operation, there is an overlapping relationship between the time window between the satellite and the earth station and the feeding time window between the satellite and the feeding antenna earth station. This overlapping relationship allows for antenna-switching operations, thereby extending the time required to establish communication links. During the planned period, the number of VTW obtained between the satellite and the ground station is limited due to the satellite flying around a fixed orbit, which also affects the success of the mission execution. How to select the appropriate VTW to arrange the link-building pairing sequence between satellites and ground stations in the feeding mode is the key to solving this problem.

\begin{figure}[htp]
	\centering
	\subfigure[scene1]{
		\includegraphics[width=0.4\textwidth]{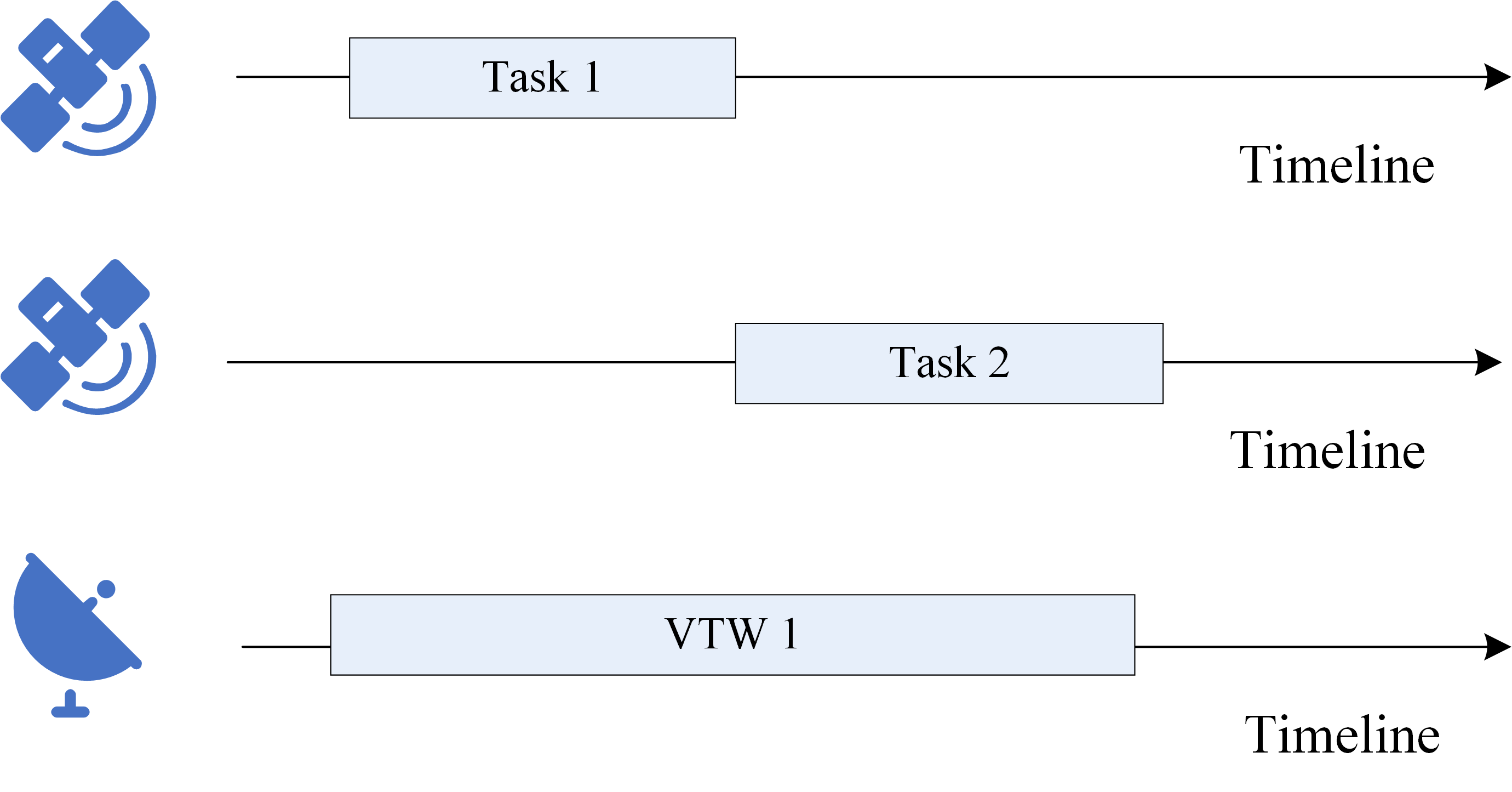}}
	\subfigure[scene2]{
		\includegraphics[width=0.4\textwidth]{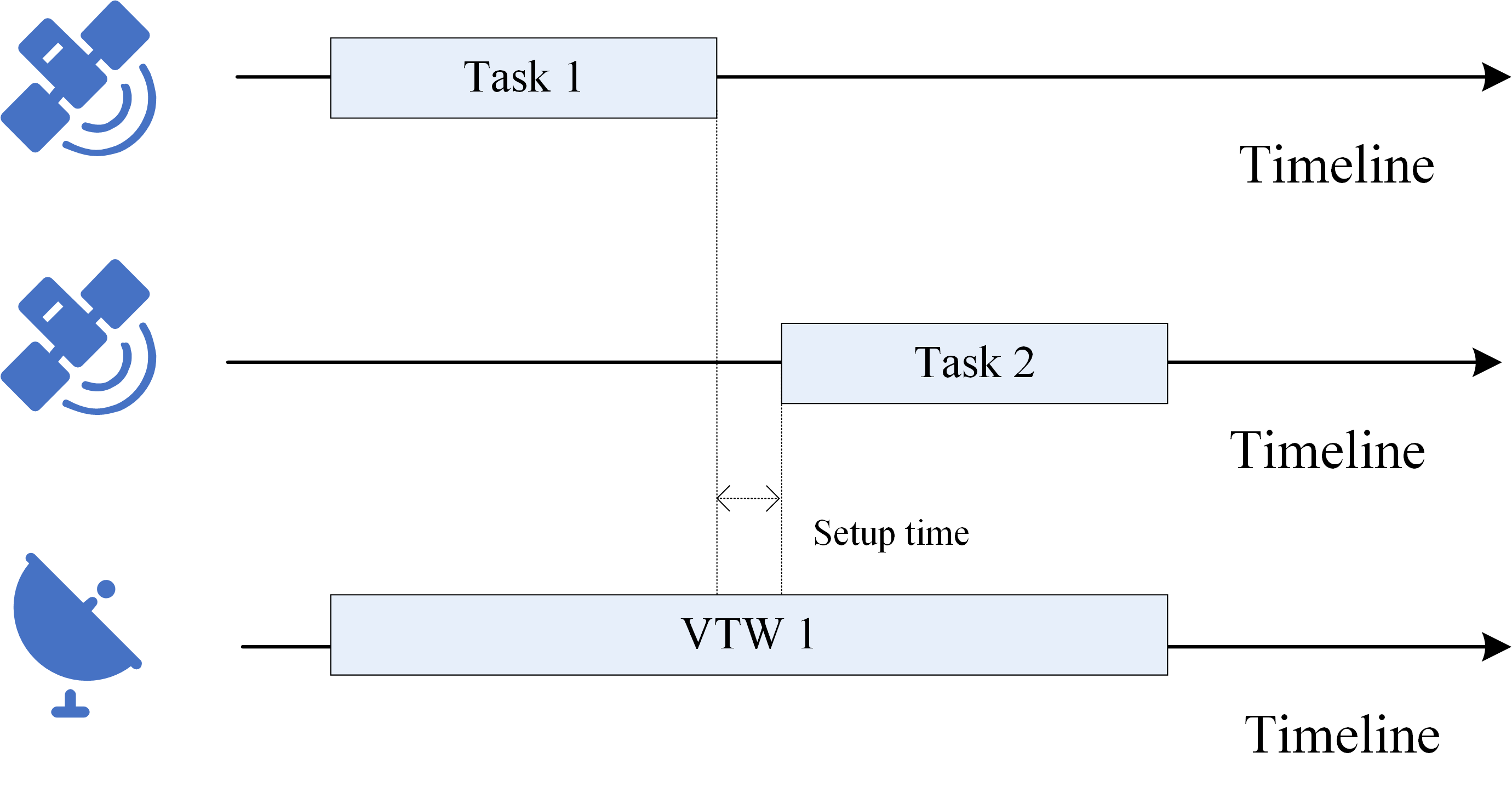}}
		
	\caption{\textcolor[rgb]{0,0,0}{Schematic Diagram of Setup Time Required for Execution between Two Tasks.}}
	\label{Schematic Diagram of Setup Time Required for Execution between Two Tasks}
\end{figure}

Each satellite has two types of antennas that can establish communication links with two types of ground stations respectively, but one satellite's antenna can only establish communication links with one type of ground station's antenna. When one of the VTW meets the conditions for task execution, the specific execution time of the task will be determined based on the equipment capabilities of the satellite and ground station. After completing one task, it takes a period of device switching before starting the next task. Usually, to improve the reliability of the link, a certain setup time is required before establishing the link, and a certain processing time is required before ending the link. As shown in Fig.\ref{Schematic Diagram of Setup Time Required for Execution between Two Tasks}, the two tasks in Fig.2(a) do not meet the switchover time requirement, resulting in the second task not being able to be executed. On the other hand, both tasks in Fig.2(b) can be successfully executed because they meet the switching time requirement of the ground station. \textcolor[rgb]{0,0,0}{In traditional mode, if the duration of a task exceeds the selected VTW, the task is not allowed to be executed, which means there is no execution sequence for the task in the current execution plan. In the SGNPFM problem studied in this article, the feed-switching mode allows the satellite to decide whether to perform feed-switching when the duration of a task exceeds the selected VTW, rather than directly canceling the task execution.} 

\begin{figure*}[htp]
	\centering
	\includegraphics[width=0.8\textwidth]{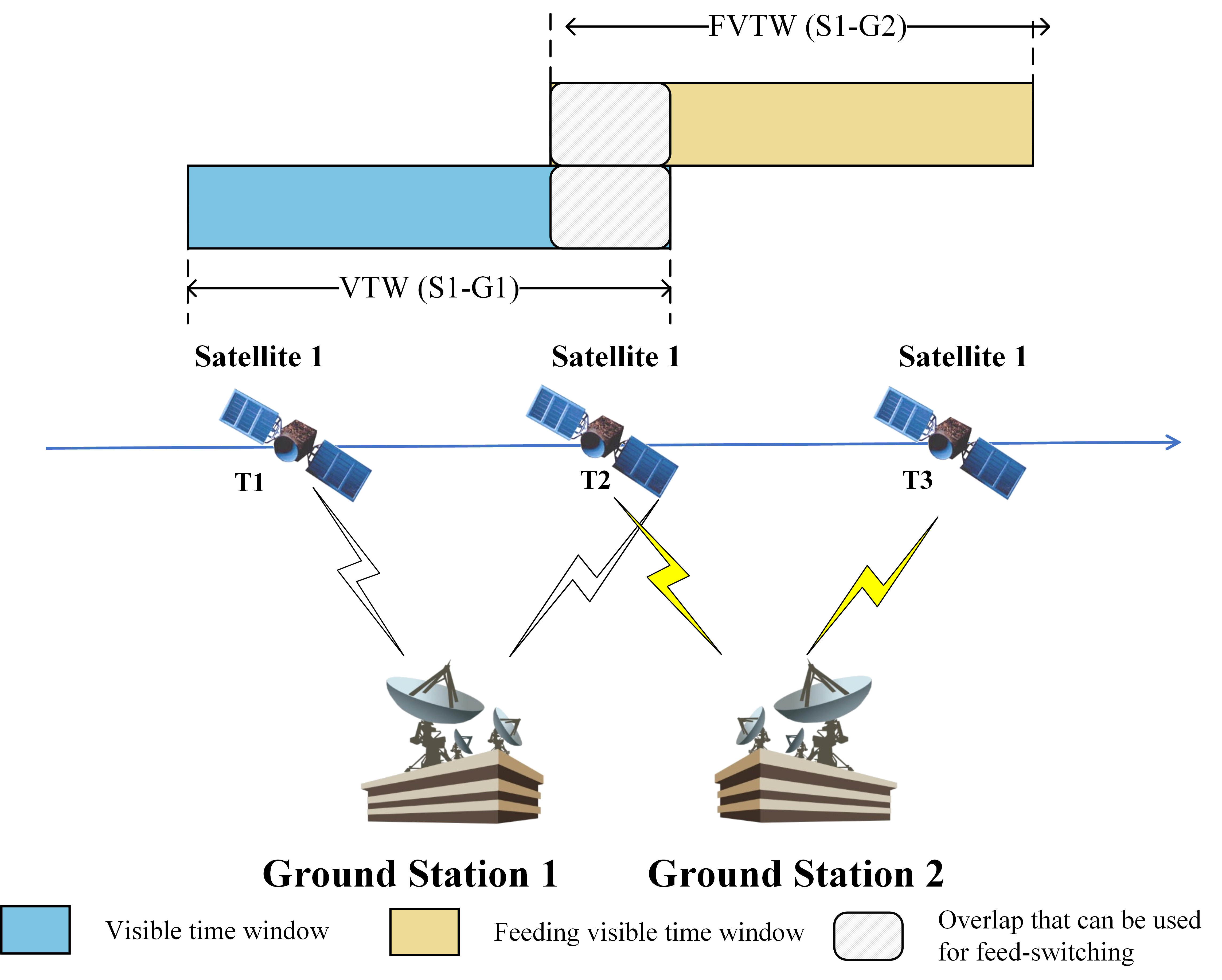}
	\caption{\textcolor[rgb]{0,0,0}{Schematic Diagram of Satellite Feed-switching.}}
	\label{Schematic Diagram of Satellite Feed-switching}
\end{figure*}

 \textcolor[rgb]{0,0,0}{In the SGNPFM problem, in addition to traditional ground stations, there are also ground stations carrying feeding antenna equipment (feeding ground stations). Compared to antennas on traditional ground stations, feeding antennas on feeding ground stations also have a certain acceptance range with satellites. The reception range between the feeding ground station and the satellite is called the Feeding Visible Time Window (FVTW). The difference between ground stations carrying feeding antennas and traditional ground stations is that feeding antennas allow for the establishment of a new connection while establishing links between satellites and other ground stations. After the successful establishment of this new link, a new FVTW constraint is established. This means that the VTW and FVTW formed by adjacent ground stations and feeding ground stations respectively overlap. A task can also be executed on FVTW. Therefore, when the selected VTW cannot meet the execution time of a task, the FVTW adjacent to and overlapping with this VTW can continue to execute the task, which satisfies the execution conditions of this task. As shown in the Fig.\ref{Schematic Diagram of Satellite Feed-switching}, the process of overlapping part between two different ground stations is illustrated. The resources for feeding ground stations are limited, and not all VTW have overlapping FVTW. Therefore, it is necessary to determine whether there are overlapping part.} 

When the duration of one task exceeds the selected VTW, the satellite will consider whether there is an overlapping part of the VTW and adopt the feed-switching working mode. \textcolor[rgb]{0,0,0}{In this mode, when the satellite establishes a link with the ground station to perform a task, it can directly switch to a new link with another ground station to continue the task without waiting time. Fig.\ref{Schematic Diagram of Satellite Feed-switching} shows the schematic diagram of satellite feed-switching. In the figure, when the satellite reaches time T2, VTW overlaps with FVTW, indicating that the satellite can establish a link with the feeding ground station at this time. Adopting this mode of working not only increases the completion rate of tasks but also reduces the time wasted in switching between tasks.} In summary, the SGNPFM problem aims to match a given series of VTWs with communication tasks, organize the task sequence, and develop a reasonable work plan to maximize the overall profit while satisfying various constraints in both regular and feed-switching modes.

\subsection{Symbols and Variables}

$S$: Satellite set, index is $i$, $S=\left\{s_i \mid i=1,2,\dots,\left |  s\right |  \right\}$. For satellite $s_i$, the following attributes are defined:

$SA^i$: Antenna set of satellite $s_i$, the number of $s_i$ is $M$, index is $m$. For antenna $sa_{m}^{i}$, the following attributes are defined:

$\alpha$: Attitude adjustment time of an antenna.

$\beta$: Minimum overlap time to complete feed-switching.

$G$: Ground station set, index is $j$, $G=\left\{g_j \mid j=1,2,\dots,\left |  g\right |  \right\}$. For satellite $g_j$, the following attributes are defined:

$GA^j$: Antenna set of ground station $g_j$, the number of $g_j$ is $N$, index is $n$. For antenna $ga_{n}^{j}$, the following attributes are defined:

$\gamma$: Minimum interval time for antenna task switching of ground station.

$T$: Task set, index is $k$, $T=\left\{t_k \mid k=1,2,\dots,\left | k\right |  \right\}$. For task $t_k$, the following attributes are defined:

$\left[est_k,let_k\right]$: \textcolor[rgb]{0,0,0}{The earliest allowable start time and latest allowable end time of task $t_k$. }

$\left[st_k,et_k\right]$: \textcolor[rgb]{0,0,0}{The start time and end time of task $t_k$.}

$p_k$: \textcolor[rgb]{0,0,0}{Profit of task $t_k$.}

$d_k$: \textcolor[rgb]{0,0,0}{Duration of task $t_k$.}

$TW$: Time window set, the time window formed by antenna $sa_{m}^{i}$ and antenna $ga_{n}^{j}$ is $TW^{ijmn}$, index is $p$. For time window ${tw}_p^{ijmn}$, the following attributes are defined:

$\left[{tws}_p^{ijmn},{twe}_p^{ijmn}\right]$: The start time and end time of the visible time window ${tw}_p^{ijmn}$.

$FTW$: Feeding time window set, the feeding time window formed by antenna $sa_{m}^{i}$ and antenna $ga_{n}^{j}$ is $FTW^{ijmn}$, index is $q$. For time window ${ftw}_q^{ijmn}$, the following attributes are defined:

$\left[{ftws}_q^{ijmn},{ftwe}_q^{ijmn}\right]$: The start time and end time of feeding time window ${ftw}_q^{ijmn}$.

$x_{kpq}^{ijmn}$: A 0-1 decision variable, which indicates whether communication task $t_k$ is executed by satellite antenna $sa_{m}^{i}$ and ground station antenna $ga_{n}^{j}$ in time window ${tw}_p^{ijmn}$ and feeding time window ${ftw}_q^{ijmn}$, if the task is executed, $x_{kpq}^{ijmn}=1$; otherwise, $x_{kpq}^{ijmn}=0$.

\subsection{Mathematical Model}
Our model for SGNPFM is based on the following assumptions:

\textbf{Assumptions:}

1. All of these link building tasks are identified before task planning begins, and there is no chance that individual tasks will be forced to be canceled due to external factors during the planning process;

2. The satellites and ground stations can remain operational at all times;

3. Each task is allowed to be executed at most once and there are no multiple or periodic executions;

4. Satellites are in proper working order to ensure the successful completion of the mission;

5. Once a task has been successfully planned, it is bound to be executed and there is no chance of it being canceled;

\hspace*{\fill}

In both regular operating and feed-switching operating modes, we aim to acquire high-quality tasks while maintaining communication link connectivity, which serves as the foundation for communication task stability. Consequently, the objective of SGNPFM is to obtain the highest profit from link-building tasks performed. The objective function is represented as follows.

\hspace*{\fill}

\textbf{Objective function:}
\begin{equation}
\label{objective}
\mathrm{Max}\sum_{m\in SA^i}\sum_{n\in GA^j}\sum_{p\in TW}\sum_{q\in FTW}\sum_{k\in T}p_k\cdot x_{kpq}^{ijmn}
\end{equation}	

\noindent \textcolor[rgb]{0,0,0}{where $p_k$ denotes profit of task $t_k$, $x_{kpq}^{ijmn}$ denotes whether communication task $t_k$ is executed by satellite $sa_{m}^{i}$ antenna  and ground station antenna $ga_{n}^{j}$ in time window $tw_{p}^{ijmn}$ and feeding time window $ftw_{q}^{ijmn}$.}  

\hspace*{\fill}

\textbf{Constraints:}
\begin{equation}
\begin{split}
	(\text{st}_k + \text{d}_k) \cdot x_{kp}^{ijmn} &\leq \text{let}_k,\forall k \in T, \, p \in TW, \, i \in S, \, j \in G,  \, m \in SA^i, \, n \in GA^j
\end{split}
\end{equation}
\begin{equation}
\begin{split}
	\mathrm{st}_{\mathrm{k}} &\geq \text{tws}_{p}^{ijmn} \cdot x_{kp}^{ijmn},\forall k \in T, \, p \in TW, \, i \in S, \, j \in G, \, m \in SA^{i}, \, n \in GA^{j}
\end{split}
\end{equation}	
\begin{equation}
\begin{split}
	\mathrm{st}_\mathrm{k} &\leq \mathrm{twe}_p^{ijmn} \cdot x_{kp}^{ijmn}, \forall k \in T, \, p \in TW, \, i \in S, \, j \in G, \, m \in SA^i, \, n \in GA^j
\end{split}
\end{equation}	
\begin{equation}
\begin{split}
	\text{est}_k &\leq \mathrm{st}_k \cdot x_{kp}^{ijmn}, \forall k \in T, \, p \in TW, \, i \in S, \, j \in G,  \, m \in SA^i, \, n \in GA^j
\end{split}
\end{equation}	

\begin{equation}
	\begin{split}
		\mathrm{let}_k &> \mathrm{st}_k \cdot x_{kp}^{ijmn}, \forall k \in T, \, p \in TW, \, i \in S, \, j \in G,  \, m \in SA^i, \, n \in GA^j
	\end{split}
\end{equation}
\begin{equation}
	\begin{aligned}
		( \mathrm{et}_{\mathrm{k}} + \alpha ) \cdot x_{kp}^{ijmn} & \leq \mathrm{st}_{\mathrm{k'}} \cdot x_{k'p}^{ijmn}, \forall k, k' \in T, \, p \in TW, \,\, i \in S, \, j \in G, \, m \in SA^i, \, n \in GA^j
	\end{aligned}
\end{equation}
\begin{equation}
	\begin{split}
		(twe_{p}^{ijmn} + \gamma) &\leq tws_{p'}^{ijmn}, \forall p, p' \in TW, \, i \in S, \, j \in G,  \, m \in SA^i, \, n \in GA^j
	\end{split}
\end{equation}

\begin{equation}
	\sum_{i\in S}\sum_{m\in SA^i}\sum_{j\in G}x_{kp}^{ijmn}\leq1,\forall j\in S,m\in SA^i,j\in G
\end{equation}	

\begin{equation}
	\sum_{i\in S}\sum_{j\in G}\sum_{n\in GA^j}x_{kp}^{ijmn}\leq1,\forall i\in S,j\in G,n\in GA^j
\end{equation}	
\begin{equation}
	\begin{split}
		(ftws_q^{ijmn} - twe_p^{ijmn}) \cdot x_{kpq}^{ijmn} &\leq \beta,\forall k \in T, \, p \in TW,  \, q \in FTW, \, m \in SA^i, \, n \in GA^j
	\end{split}
\end{equation}

\begin{equation}
	\begin{split}
		twe_{p}^{ijmn} &\leq (\mathrm{st}_{\mathrm{k}} + \mathrm{d}_{\mathrm{k}}) \cdot x_{kpq}^{ijmn} \leq ftwe_{q}^{ijmn},\forall k \in T, \, p \in TW, \, q \in FTW, \, i \in S, \, j \in G, \, m \in SA^i, \, n \in GA^j
	\end{split}
\end{equation}
\begin{equation}
	\begin{split}
		\sum_{k \in T} \sum_{p \in TW} \sum_{q \in FTW} x_{kpq}^{ijmn} &\leq 1, \forall k \in T, \, p \in TW, \, q \in FTW,\, i \in S, \, j \in G, \, m \in SA^i, \, n \in GA^j
	\end{split}
\end{equation}

\begin{equation}
	x_{kpq}^{ijmn}\in\{0,1\}
\end{equation}	

Constraint (2) indicates that the start time of the task is earlier than the completion time. Constraint (3) indicates that the task needs to start within the visible time window. Constraint (4) indicates that the task needs to be completed within the visible time window. Constraint (5) indicates that the task needs to start after the earliest allowable start time of the task. Constraint (6) indicates that the task needs to be completed before the latest allowable time of the task. Constraint (7) indicates that after a completing a task, the satellite needs to meet the minimum interval time requirement to execute the next task. Constraint (8) indicates that after completing a task, the ground station needs to meet the minimum interval time requirement for antenna task switching to execute the next task. Constraint (9) indicates that a satellite can only have one antenna to build a communication link with a ground station at a moment in time. Constraint (10) indicates that a ground station have one antenna to build a communication link with a satellite antenna at a moment in time. Constraint (11) indicates that the overlap of the two visible time windows in which satellite performs the feed-switching operation is to exceed the minimum overlap time. Constraint (12) indicates that when a satellite performs the feed-switching operation, the task needs to be completed within the feeding time window. Constraint (13) indicates that one task can be executed at most once. Constraint (14) indicates that the value range of the decision variable.

\section{The Proposed Method}
\label{The proposed method}
To solve the SGNPFM problem, a genetic optimization algorithm based on similarity evaluation is proposed as a new idea to solve the problem. The enhanced algorithm encompasses the following three innovative aspects: 1) An improved genetic algorithm based on similarity mechanism is designed to dynamically determine the processing mode of offspring individuals, which includes operations for population selection and population mutation based on similarity; 2) Based on SGNPFM, we consider the feed-switching mode and specifically design a new task scheduling algorithm task scheduling method with feed switching mode (TSMFS); 3) A series of improvement strategies based on the knowledge and similarity mechanisms in the problem are proposed, including heuristic initialization methods, adaptive crossover strategies, which guide the algorithms to search efficiently. This section will introduce the overall framework of DSGA and the improvement methods used in the algorithm.

\subsection{Distance Similarity-based Genetic Optimization Algorithm (DSGA)}
Distance similarity-based genetic optimization algorithm (DSGA) is the core algorithm for solving the SGNPFM problem. Genetic optimization algorithms are improved based on the analysis of problem characteristics and task data features. This section will introduce the process of the genetic optimization algorithm based on distance similarity, the key steps in the algorithm and the adaptive crossover strategy.

\subsubsection{Algorithm Overall Process}
\begin{figure*}[htp]
	\centering
	\includegraphics[width=0.75\textwidth]{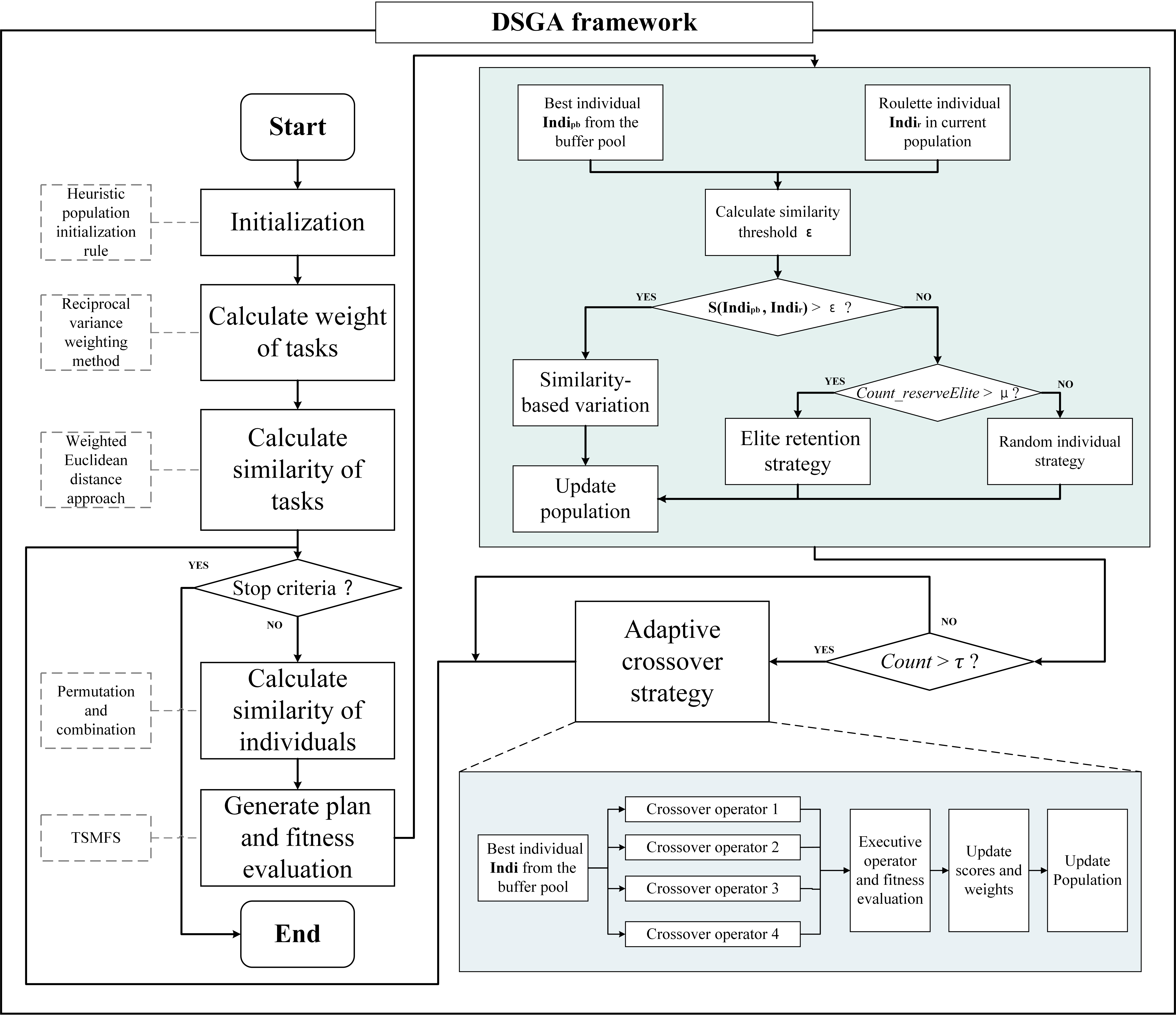}
	\caption{\textcolor[rgb]{0,0,0}{The Framework of DSGA.}}
	\label{The Framework of DSGA}
\end{figure*}

The conventional genetic algorithm, as an optimization technique, exhibits certain limitations including slow convergence speed and susceptibility to local optima. We improve the traditional genetic algorithm framework and propose a distance similarity-based genetic optimization algorithm (DSGA). \textcolor[rgb]{0,0,0}{DSGA fully uses the task's attribute data characteristics which serve as the evaluation basis for similarity. The mechanism of distance similarity has been introduced by DSGA, which evaluates the similarity between tasks and individuals and proposes an adaptive crossover strategy based on similarity mechanism to efficiently and targetedly search the population. The overall process of DSGA is shown in the Fig.\ref{The Framework of DSGA}}

\textcolor[rgb]{0,0,0}{Firstly, DSGA adopts heuristic initialization rules to generate high-quality populations. A high-quality population can gain more advantages and improve search efficiency in the early stages of algorithm iteration. Subsequently, the attribute weights of tasks and the similarity between tasks need to be calculated. This part is a prerequisite for the algorithm to perform, as the selection and crossover operations of the population are based on similarity mechanisms. Then, DSGA calculates the similarity of individuals in iterations to select different individuals. The similarity between two individuals determines the strategy adopted by the population for selection. Finally, when the number of algorithm iterations reaches a certain threshold, an adaptive crossover strategy is implemented. This strategy dynamically uses different operators to perturb individuals based on the scores assigned at different times, improving the algorithm's local search ability.}

\textcolor[rgb]{0,0,0}{The pseudo code of DSGA is shown in Algorithm \ref{a:dsga}. A heuristic population initialization strategy is employed to generate initial populations (Line 2). The similarity matrix results between tasks are obtained using a similarity calculation method based on weighted Euclidean distance (Line 3). A similarity generation method based on permutation and combination is utilized to derive the similarity matrix results between individuals (Line 5). TSMFS is employed for decoding and fitness value computation based on task profitability (Line 6). To enhance convergence, an elite retention strategy involving a buffer pool is implemented for storing the best individuals (Line 11). Furthermore, an adaptive crossover strategy based on improving offspring through enhanced similarity is applied depending on different thresholds and individuals (Line 18 and Line 32).}

\begin{algorithm}[htbp]
\label{a:dsga}
	\caption{Distance similarity-based genetic optimization algorithm}
	\small
	\LinesNumbered %要求显示行号
	\KwIn {Population size $N$, Task set $T$, Time window set $TW$, Feeding time window set $FTW$, Buffer pool $pool$, Thre $\varepsilon, \mu, \tau, \varphi$}
	\KwOut {$ Solution $}%输出
	Initialize algorithm parameters\;
	$ P_0 \leftarrow $	Generate initial population by Heuristic population initialization method in Section \ref{Heuristic population initialization method}\;
	$ T_s \leftarrow $ Calculate similarity between tasks // \textcolor[rgb]{0,0,0}{Use weighted Euclidean Distance}\;
	\While{termination criterion is not met}{
		$ I_s \leftarrow $ Calculate similarity between individuals $ \left(T_s,P_0\right) $ // 
 \textcolor[rgb]{0,0,0}{Based on permutation and combination methon}\;
		$ R \leftarrow $ Generate scheduling plan by TSMFS in Section \ref{tsmfs}$ \left(T,TW,FTW,P_0\right) $\;
		$ F_p \leftarrow $ Fitness evaluation$ \left( R \right) $\;
		\eIf{$ f_{local} > f_{global}$}
		{
			$f_{global} \leftarrow f_{local}$\;
			$indi_{global} \leftarrow indi_{local} $\;
			$pool$.append $\left(indi_{global},f_{gobal}\right) $\;
		}
		{
			$count\_reserveElite \leftarrow  count\_reserveElite  +1 $\;
		}
             $indi_r \leftarrow$ Select individual by roulette wheel method\;
		\For{$i \leftarrow 1$ to $N_p$}
		{
			$indi_{i} \leftarrow $ Select the current individual $i$ in the population\;
			
			\eIf{ Find individual similarity  $ \left(indi_{i},indi_r,I_s\right) > \varepsilon $}
			{
				$indi_{i'} \leftarrow $ Use adaptive crossover strategy in Section \ref{Adaptive crossover strategy}\;
				$P'\leftarrow $ Append population $ \left(indi_i'\right)$\;
			}
			{	
				\eIf{ $count\_reserveElite > \mu $}
				{
                    $ indi_{pb} \leftarrow $ Find the best individual of current pool\;
					$ P' \leftarrow $ Append population $ \left(indi_{pb}\right)$\;
					Reset $ count\_reserveElite $\; 
				}
				{
					$ P' \leftarrow $ Append population $ \left(indi_{r}\right)$\;
				}
			}
		}
		Update population\;
		\textcolor[rgb]{0,0,0}{ $count$} $ \leftarrow $ \textcolor[rgb]{0,0,0}{$count $} $ +1 $\;
		\If{ \textcolor[rgb]{0,0,0}{count} $ > \tau$}
		{
			\eIf{$ rand\left(\right) < \varphi$}
			{
				$ P \leftarrow$ Randomly generate a new individual\;
			}
			{
				$ P \leftarrow$ Find the best individual of pool // \textcolor[rgb]{0,0,0}{Use adaptive crossover strategy in Section \ref{Adaptive crossover strategy}}\;
			}
			Reset count\;
		}			
		
	}
\end{algorithm}

\subsubsection{Encoding and Decoding}

\begin{figure}[htp]
	\centering
	\includegraphics[width=0.3\textwidth]{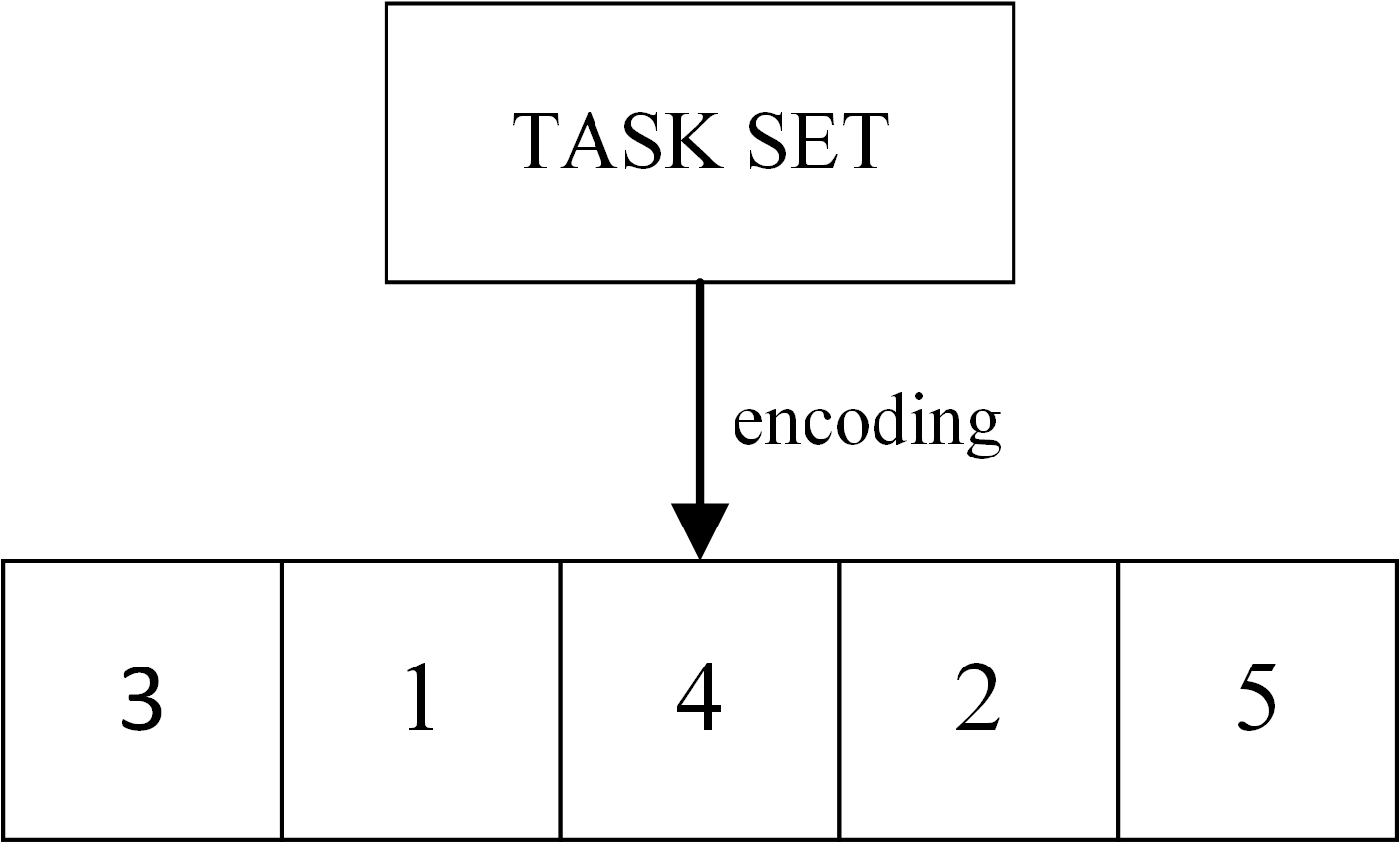}
	\caption{\textcolor[rgb]{0,0,0}{An Example of Individual Encoding.}}
	\label{An Example of Individual Encoding}
\end{figure}

The encoding is crucial for genetic algorithm and directly affects the calculation of individual fitness and the generation of programs. In the DSGA, we use an integer number encoding method to generate sequences of link building tasks into a gene sequence. Each gene represents a task to be executed. For ease of understanding, a simple example is used here for visualization. As shown in Fig.\ref{An Example of Individual Encoding}, the construction of a communication link utilizing a randomized approach to generate a gene fragment can be achieved through a sequence of six tasks. If there is a gene fragment of “3 1 4 2 5”, it means that the first task in this part of the gene fragment is a communication task numbered “3” and the second task is communication task numbered “1”, followed by the tasks numbered “4, 2, 5”.

Decoding is the process of generating a task execution scheme, and in SGNPFM the fitness is to be calculated based on the task execution scheme under the feed-switching constraints. In this paper, TSMFS is used to complete the decoding process in the DSGA for solving the SGNPFM problem. TSMFS is introduced in Section \ref{tsmfs}. When a gene sequence is given, the method will complete the corresponding task execution scheme generation one by one in the order from front to back. If the tasks have been arranged, the decoding process is completed.

\subsubsection{Heuristic Population Initialization Method}
\label{Heuristic population initialization method}
The quality of the initial population has a great impact on the subsequent optimization, so it is necessary to initialize the population for optimization. If all random generation methods are used, the search efficiency is obviously not high. To improve the search efficiency at the beginning of the algorithm iteration, we use a heuristic population initialization method to construct the initial population by using the knowledge of the classical satellite scheduling problem. This new population initialization method combines the number of heuristic rules to reduce multiple ineffective searches of the genetic algorithm.

For initial population generation before optimization, we use three rules for heuristic population initialization. Among them, each rule is designed to generate 20\% of the population, and the remaining 40\% of the individuals in the population are completely random. Fig.\ref{Formation Structure of Initial Population} shows the formation structure of the initial population. The following are the three heuristic population initialization rules:

\begin{figure}[htp]
	\centering
	\includegraphics[width=0.6\textwidth]{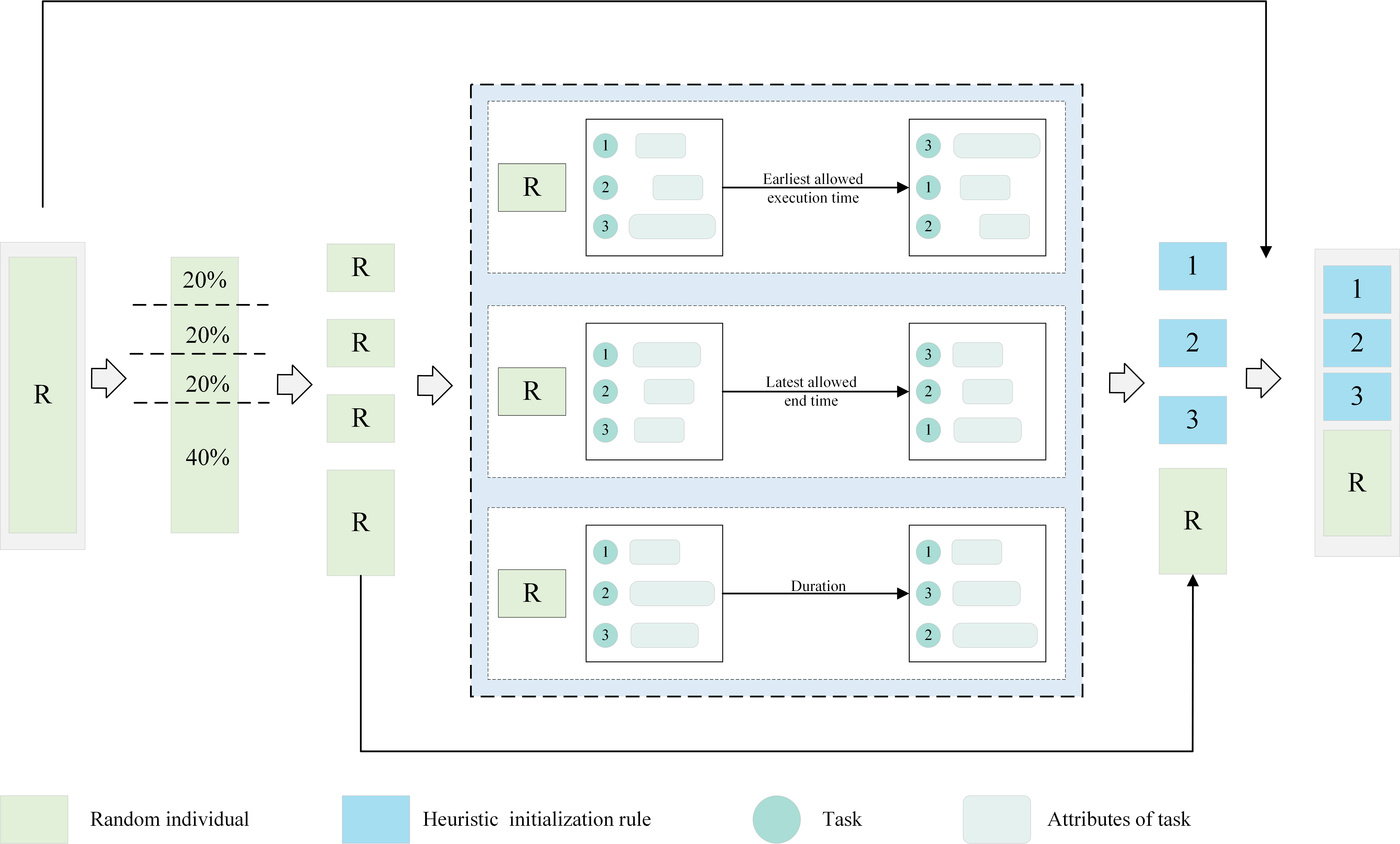}
	\caption{\textcolor[rgb]{0,0,0}{Formation Structure of Initial Population.}}
	\label{Formation Structure of Initial Population}
\end{figure}

\textbf{Heuristic population Initialization Rule 1 (HIR1):} An individual is first randomly generated and then sorted from smallest to largest by the earliest allowed execution time of the task.

\textbf{Heuristic population Initialization Rule 2 (HIR2):} An individual is first randomly generated and then sorted from smallest to largest by the latest allowed end time of the task.

\textbf{Heuristic population Initialization Rule 3 (HIR3):} An individual is first randomly generated and then sorted from smallest to largest by the task duration.

The combination of three heuristic population initialization rules and a randomization method not only considers the principle of population diversity but also ensures the acquisition of high-quality initial solutions during the population initialization process.

\subsubsection{Similarity Calculating Method}
\begin{figure}[htp]
	\centering
	\includegraphics[width=0.35\textwidth]{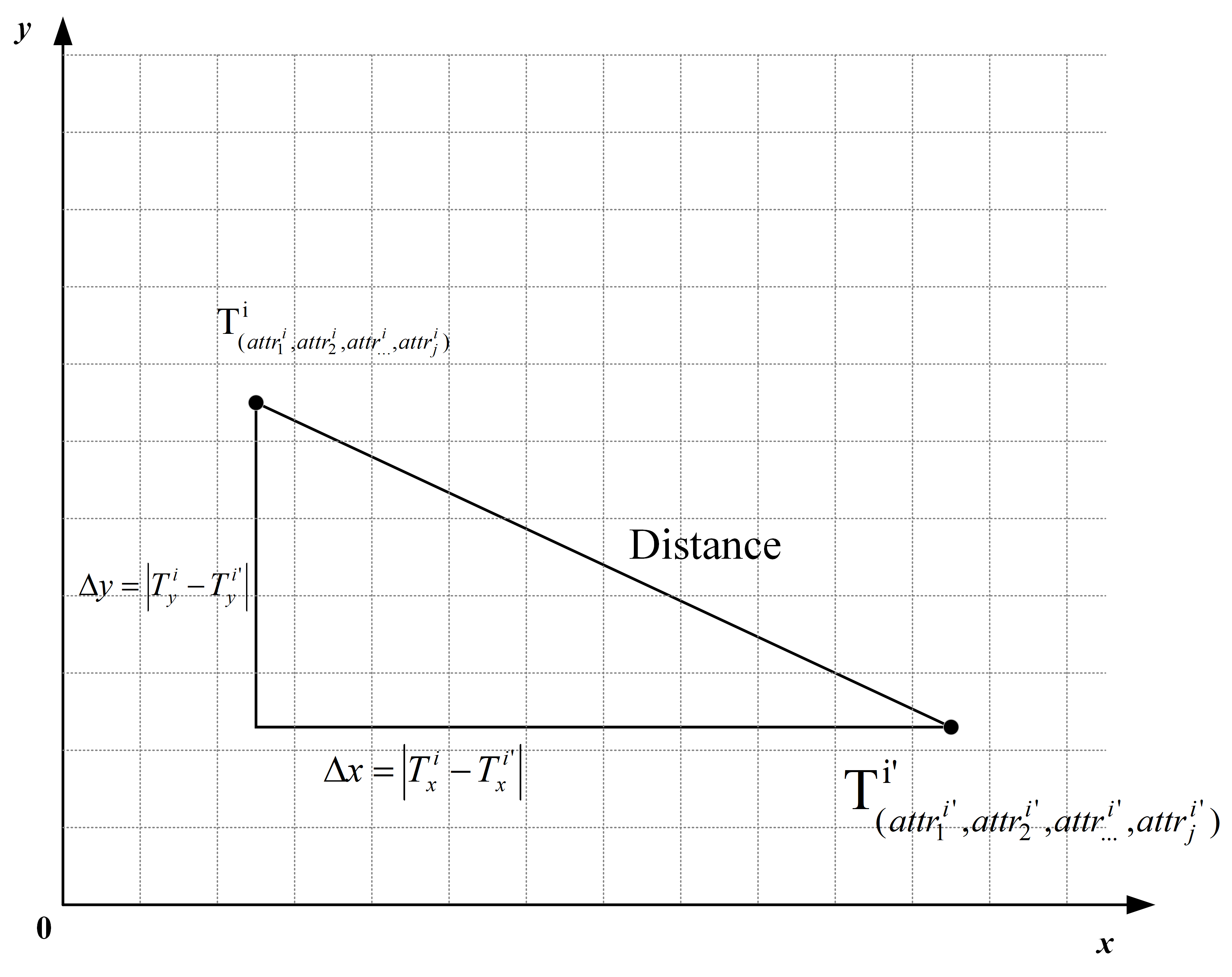}
	\caption{\textcolor[rgb]{0,0,0}{Schematic Diagram of The Distance to Task Evaluation.}}
	\label{Schematic Diagram of The Distance to Task Evaluation}
\end{figure}

Similarity is a metric used to measure the degree of similarity between two objects. The attribute states of the communication tasks in SGNPFM, such as arrival state, duration state, gain state, etc., can exert an influence on the final execution order of these tasks. To make full use of the state characteristics of the task, we use the distance similarity evaluation method and integrate it into the genetic algorithm framework. Distance is used to indicate the similarity between tasks. Specifically, distance is used for attribute differences between different communication tasks in the SGNPFM problem. By calculating the distance between tasks, we can evaluate their similarity. There are two advantages of using this method. On the one hand, the similarity evaluation method can consider the relationship between communication tasks and better evaluate the similarity between individuals within the population, so as to guide the genetic algorithm to explore more effectively in the search space; On the other hand, the method can utilize the similarity information to enhance the preservation and selection of dominant individuals, so that more similar dominant individuals can be concentrated in a population to improve the quality of the solution and the convergence speed. Fig.\ref{Schematic Diagram of The Distance to Task Evaluation} is a schematic diagram of the distance to task evaluation. \textcolor[rgb]{0,0,0}{In Fig.\ref{Schematic Diagram of The Distance to Task Evaluation}, each communication task has distinct attribute states. For instance, two tasks with similar arrival time may have different duration or profit, or two tasks with the same profit may have significantly different arrival time. Additionally, the differences in these attribute states are determined by variations in data size, making the complexity of processing such data relatively low. Based on these characteristics, we use the difference in data distance to distinguish between tasks. To prioritize tasks with higher cost-effectiveness, we adopt a similarity evaluation method based on distance differences. Thus, we establish a connection between distance and similarity evaluation.}

Among various similarity calculation methods, the Euclidean distance method demonstrates superior performance in quantifying the similarity between tasks as it measures the spatial distance between two vectors. The Euclidean distance is a fundamental concept in geometric measurement, representing the shortest path (straight line distance) between two points. This intuitive physical interpretation renders the Euclidean distance highly comprehensible and explicable. Moreover, it finds extensive application across diverse domains owing to its suitability for real vector space data. However, the traditional Euclidean distance method does not consider the actual data distribution. It only calculates the length to determine the similarity between vectors, which can lead to inaccurate similarity measurements between tasks. In this paper, we aim to determine the similarity that depends not only on the similarity of the sample vectors but also on the intrinsic nature of the data itself.

To improve the accuracy of the similarity, we used a Weighted Euclidean distance approach to calculate the similarity of the tasks. This method focuses more on the effect of the object's attributes on the distance than the traditional Euclidean distance method and it gives more attention to the attributes that are more important to the problem we are studying through the strategy of additional weights to obtain higher accuracy. This method will be introduced in detail.

First, we quantify the attribute data of the task into a spatial vector, and obtain the weight value of each attribute through the \textbf{Reciprocal Variance Weighting (RVW)} method, which will be used to evaluate the importance of the task's attributes to the similarity result. The higher the value of the weights, the greater the degree of variation in the task attributes and the greater the result of the impact on the similarity. \textcolor[rgb]{0,0,0}{RVW is defined as the ratio of the standard deviation to the mean of an attribute, used to reflect the degree of dispersion of the attribute.  The reason for using this method is that the importance of different attribute features in evaluating similarity varies. Through RVW method, the contribution of attributes with significant changes in similarity calculation to distance will also increase accordingly, helping the algorithm to more accurately calculate the differences between samples.} The calculation formula is as follows.
\begin{equation}
	\overline{x}=\frac{1}{n}\sum_{i=1}^nX_i
\end{equation}
\begin{equation}
	S_x=\sqrt{\frac{1}{n-1}\sum_{i=1}^n{(X_i-\overline{x})^2}}
\end{equation}

\noindent where $\bar{x}$ represents the average value of a task attribute state, $X_i$ represents the current data of a task attribute state. $S_x$ is the variance of task attribute state $x$, $x_1,x_2,\cdots ,x_n$ represents $n$ data of an arbitrary attribute service state $x$. 

After obtaining the variance, \textcolor[rgb]{0,0,0}{we calculate the reciprocal coefficient of variance by dividing the variance by the average value. The proportion of the sum of these coefficients gives us the required weight value.} For convenience, for the selected attributes in the task, we use $1,2,\cdots,m$ to represent all $m$ attributes of the task. The calculation formula of the reciprocal coefficient of variance and weight is as follows.

\begin{equation}
	y_x=\frac{S_x}{\bar{x}}
\end{equation}
\begin{equation}
	w_i=\frac{y_i}{\sum_{i=1}^my_i}
\end{equation}

\noindent where $y_x$ is the reciprocal coefficient of variance of task attribute state $x$, $w_i$ is the similarity weight of the task attribute state $i$. 

The weighted Euclidean distance of two tasks can be calculated by incorporating the obtained weight values into the calculation process, representing the vector distance of their corresponding attribute states. To ensure a bounded range, a mathematical transformation is applied to the reciprocal plus one of the distance result. This yields a more precise similarity measure ranging from 0 to 1, where higher values indicate greater degrees of similarity between the tasks. The formulas for calculating weighted Euclidean distance and task similarity are provided below.
\begin{equation}
	\label{f:d}
	d_{(x,y)}=\sqrt{\sum_{i=1}^mw_i\cdot(x_i-y_i)^2}
\end{equation}
\begin{equation}
	\label{f:s}
	s_{(x,y)}=\frac{1}{1+d_{(x,y)}}
\end{equation}

\noindent where $d_{(x,y)}$ is the weighted Euclidean distance between task $x$ and task $x$, $s_{(x,y)}$ is the similarity between task $x$ and task $x$. 

The similarity between two tasks is calculated and stored in a similarity matrix to enhance search efficiency. This matrix provides essential information on the degree of similarity among individuals during iteration and guides the direction of population optimization. The pseudo-code for the task similarity method is shown in Algorithm \ref{a:similarty}.

\begin{algorithm}[htbp]
\label{a:similarty}
	\caption{Similarity of tasks calculating method}
	\small
	\LinesNumbered %要求显示行号
	\KwIn {Task Set $ T $}
	\KwOut {Task similartity $ T_s $}%输出
	Initialize algorithm parameters\;
	$ T_w \leftarrow $ Calculate the weight by the reciprocal of variance weighting method\;
	\ForEach{$ task_i $ in the order}
	{
		$\{attr_{1}^{i},attr_{2}^{i},\ldots,attr_{m}^{i}\} \leftarrow $ Get all attribute states of the current task $ i $\;
		\ForEach{$ task_j $ in the order}
		{	
			$\{attr_{1}^{j},attr_{2}^{j},\ldots,attr_{m}^{j}\} \leftarrow $ Get all attribute states of the current task $ j $\;
			$ d_{i,j} \leftarrow $ Calculate the Weighted Euclidean Distance of tasks by Eq.\eqref{f:d}\;
			$ s_{i,j} \leftarrow $ Calculate the similarity by Eq.\eqref{f:s}\;
			$ T_{s(i,j)} \leftarrow $ Add per similarity to task similarty matrix $\left( s_{i,j} \right)$\; 
		}
	}
\end{algorithm}

After obtaining the similarity matrix of the task, it is also necessary to calculate the pairwise similarities between individuals. In SGNPFM, individuals within a population represent sequences of tasks, and the position of each task within an individual significantly influences the outcomes. Considering the sensitivity of task location, we employed an individual similarity calculation method based on permutation and combination. \textcolor[rgb]{0,0,0}{The main idea of this method is to calculate the average task similarity between two individuals at each position.} Firstly, we unfold the sequences of two tasks. Then, we compute the sum of their corresponding task similarities iteratively. Finally, this sum is divided by the number of task sequences to derive the similarity between two individuals. For example, there are two individuals A and B whose task sequences are $A=\left\{a_1,a_2,\cdot,a_n\right\}$ and $B=\left\{b_1,b_2,\cdot,b_n\right\}$, then the similarity of $A$ and $B$ is calculated as follows.
\begin{equation}
	Indi_{S(A,B)}=\frac{\sum_{i=1}^nT_s(a_i,b_i)}n
\end{equation}

\noindent where $Indi_{S(A,B)}$ is the similarity between individual $A$ and individual $B$, $T_s$ is the task similarity matrix. The larger values of $Indi_{S(A,B)}$, the greater the similarity between individuals, and the more similar individuals are.

After obtaining the information of individual similarity, we can use this information to determine whether two individuals are similar. There is a feature of our individual similarity calculation method using permutations and combinations that as the population iterates, the degree of similarity of the individuals increases, and the similarity value between them becomes larger and larger in the interval. To address algorithmic complexity, we adopt the average individual similarity value of each generation as a selection threshold for the population, gradually reducing it with subsequent iterations. According to the average threshold of individual similarity in each generation, DSGA judges that the two tasks are similar. If their similarity value exceeds the threshold, it is determined that the two tasks are similar; otherwise, they are considered dissimilar. The formula employed to lower this threshold is presented below.
\begin{equation}
	f(x)=ave_x-std_x(1-e^{\frac{-\ln2(x-1)}{m-1}}) , x=1,2,\ldots,m
\end{equation}

\noindent where $x$ represents each iteration number of the population, $m$ represents the maximum iteration number, $f(x)$  represents the threshold selection function of the population in each iteration, $ave_x$ represents the average similarity value of individuals in the $x$ generation, $std_x$ represents the average similarity standard deviation of individuals in the $x$ generation. 

When the maximum number of iterations is reached, the lowest point of the threshold is approximated as $ave_x-std_x$. The purpose of this design is, on the one hand, to provide the population with a broader range of evolutionary choices during the initial iterations of the algorithm, and on the other hand, to enhance the convergence rate in subsequent iterations by concentrating individuals with superior qualities within the population.

\begin{figure*}[htp]
	\centering
	\includegraphics[width=0.6\textwidth]{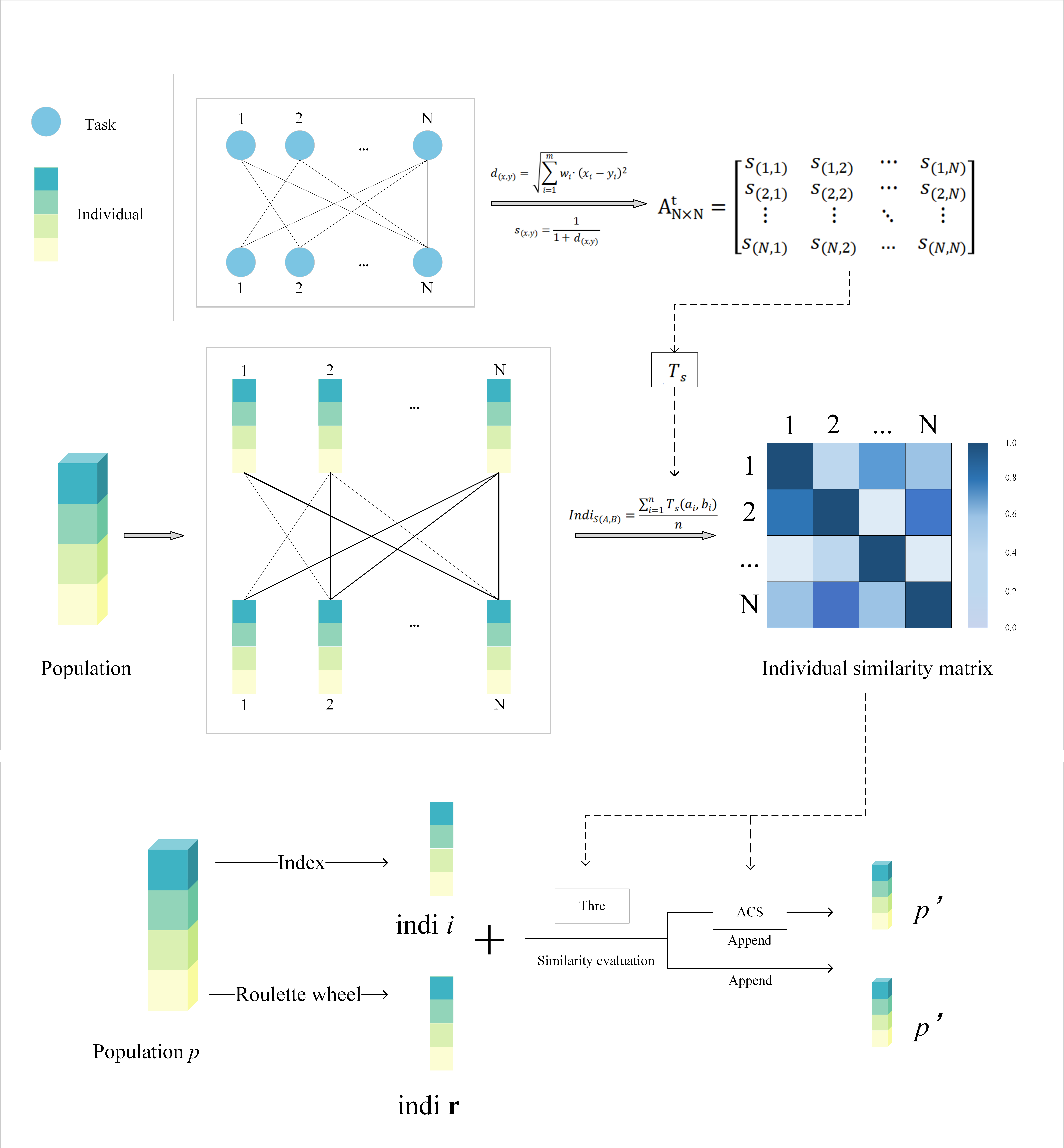}
	\caption{\textcolor[rgb]{0,0,0}{Schematic Diagram Showing The DSGA Similarity Evaluation.}}
	\label{Schematic Diagram Showing The DSGA Similarity Evaluation}
\end{figure*}

Fig.\ref{Schematic Diagram Showing The DSGA Similarity Evaluation} is a schematic diagram showing the similarity evaluation of DSGA. The upper part of the picture represents the two methods for specifically obtaining the similarity matrix of a task and the similarity matrix of an individual, as well as the connection between them. \textcolor[rgb]{0,0,0}{The meaning of $A_{N \times N}^{t}$ is the similarity matrix of tasks. $A$ represents “Array”, $t$ represents “Task”, and $N \times N$ represents the comparison between N tasks. The lower part of the picture represents the process of how the similarity evaluation method allows the selection of similar offspring individuals into the next population.  "Thre" is an abbreviation for threshold, which represents the criteria for evaluating whether two individuals are similar in the process of population selection. Additionally, the \textbf{Adaptive Crossover Strategy (ACS)} also uses the mechanism of similarity to enhance the effectiveness of the crossover operator.}

\subsubsection{Selection}
\textcolor[rgb]{0,0,0}{After obtaining the similarity matrix of all individuals and the threshold of their similarity, we can determine whether two individuals are similar or not during the selection process. }

Selection serves as the fundamental premise for population evolution in genetic algorithms. The selection operation should effectively capture the performance differences among individuals within the population. Previous studies have proposed various methods for selection operations, including roulette wheel selection, ranked selection, tournament selection, and random sampling. Among these methods, \textbf{Roulette Wheel Selection (RWS)} stands out as a widely employed mechanism. RWS involves mapping each individual's fitness value to a proportional sector on a roulette wheel. The size of each sector corresponds to the individual's fitness level. Subsequently, an individual is selected by randomly determining which sector is indicated by a pointer.

The advantages of RWS lie in its simplicity and effectiveness, making it a suitable method for reflecting the fitness differences between individuals. This enables individuals with higher fitness levels to have a greater probability of selection, thereby enhancing the overall population's fitness. Considering these benefits, we employed the RWS method for individual selection to facilitate an evolutionary process that favors the preferential selection of better-performing individuals, progressively optimizing the solution. In DSGA, an individual selected using the RWS method undergoes a mutation operation based on similarity mechanism. The equation used for selecting individuals through the RWS method is as follows.
\begin{equation}
	p_i=\frac{fit_i}{\sum_{i\in N_p}fit_i}
\end{equation}

\noindent where $p_i$ represents the fitness function value, $N$ represents the number of individuals, $p_i$ represents the probability of an individual being selected.

\textcolor[rgb]{0,0,0}{As shown in the Fig.\ref{Schematic Diagram Showing The DSGA Similarity Evaluation}, in DSGA, there are two ways for the algorithm to select individuals in each generation. One approach is to use RWS method to select an individual as the benchmark individual before iterating through the population. When this benchmark individual is selected, it serves as the basis for similarity evaluation with other individuals. The advantage of using RWS is that the benchmark individual not only has a greater possibility of becoming the optimal individual, but also has a certain possibility of becoming the individual with the best potential, which means there is more diversity. Another approach is to simply iterate through the population to obtain the current individual. The advantage of this selection method is to reduce the complexity of the selection, while also being able to explore all individuals and obtain all opportunities for comparison.}

\subsubsection{Fitness Evaluation}
Fitness calculation plays a crucial role in genetic algorithms, as it directly influences the likelihood of an individual being selected for genetic operations. In the DSGA, the performance of individuals in each population is evaluated using a fitness function. The objective function is used as the fitness function, and the fitness value of an individual is calculated using the Eq. \eqref{objective}.

\subsubsection{Adaptive Crossover Strategy}
\label{Adaptive crossover strategy}
In DSGA, we have implemented an adaptive crossover strategy consisting of two classes of operators: diverse random crossovers and proposed crossovers based on similarity mechanisms. Each class includes two types of operators, resulting in a total of four distinct operators. The following section provides a brief description of these four operator types.

\begin{figure}[htp]
	\centering
	\includegraphics[width=0.4\textwidth]{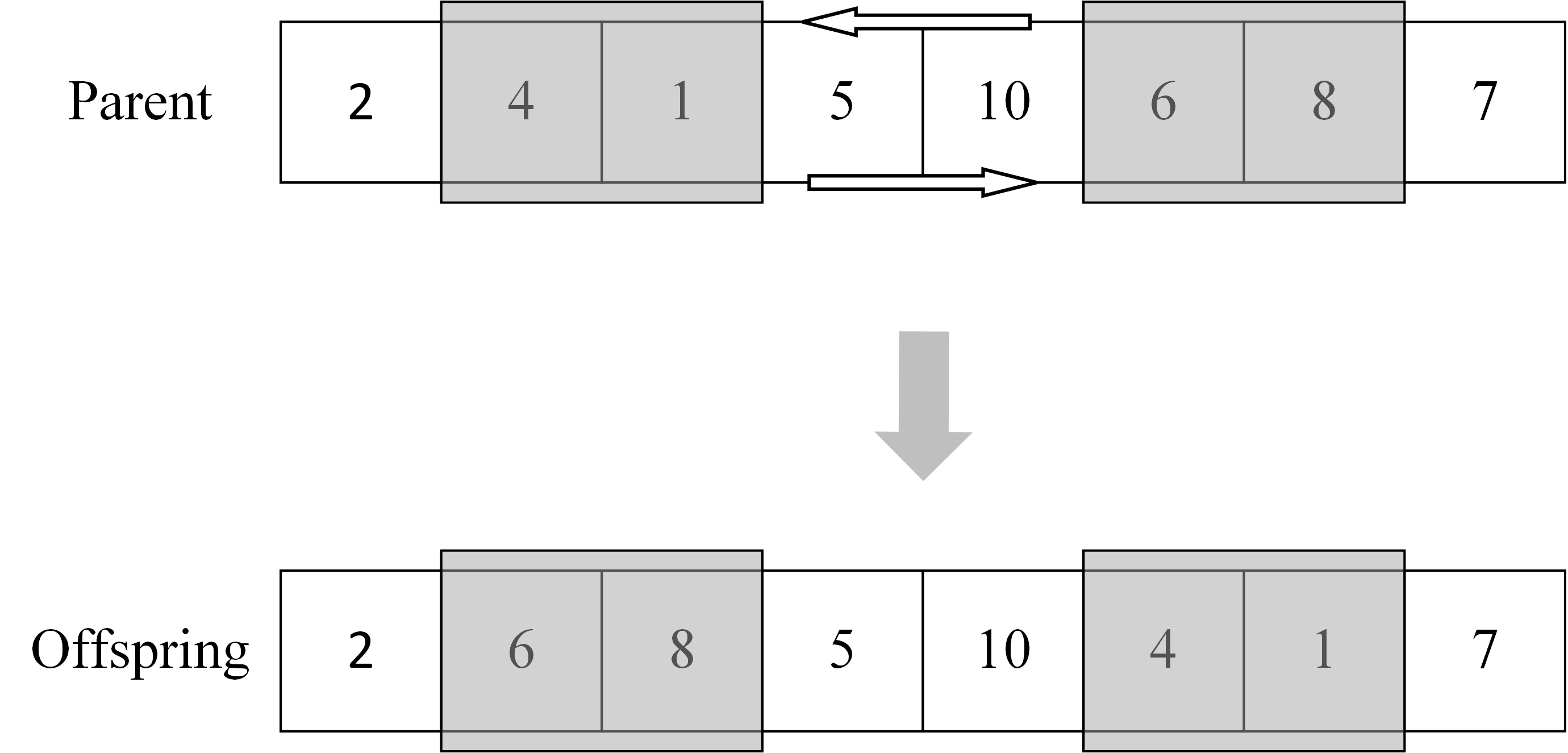}
	\caption{\textcolor[rgb]{0,0,0}{Schematic Diagram of Random Crossover Operator 1.}}
	\label{Schematic Diagram of Random Crossover Operator 1}
\end{figure}

\textbf{Random crossover operator 1}: As shown in Fig.\ref{Schematic Diagram of Random Crossover Operator 1}, randomly select two segments from a task sequence segment and swap their positions to form a new segment.

\begin{figure}[htp]
	\centering
	\includegraphics[width=0.4\textwidth]{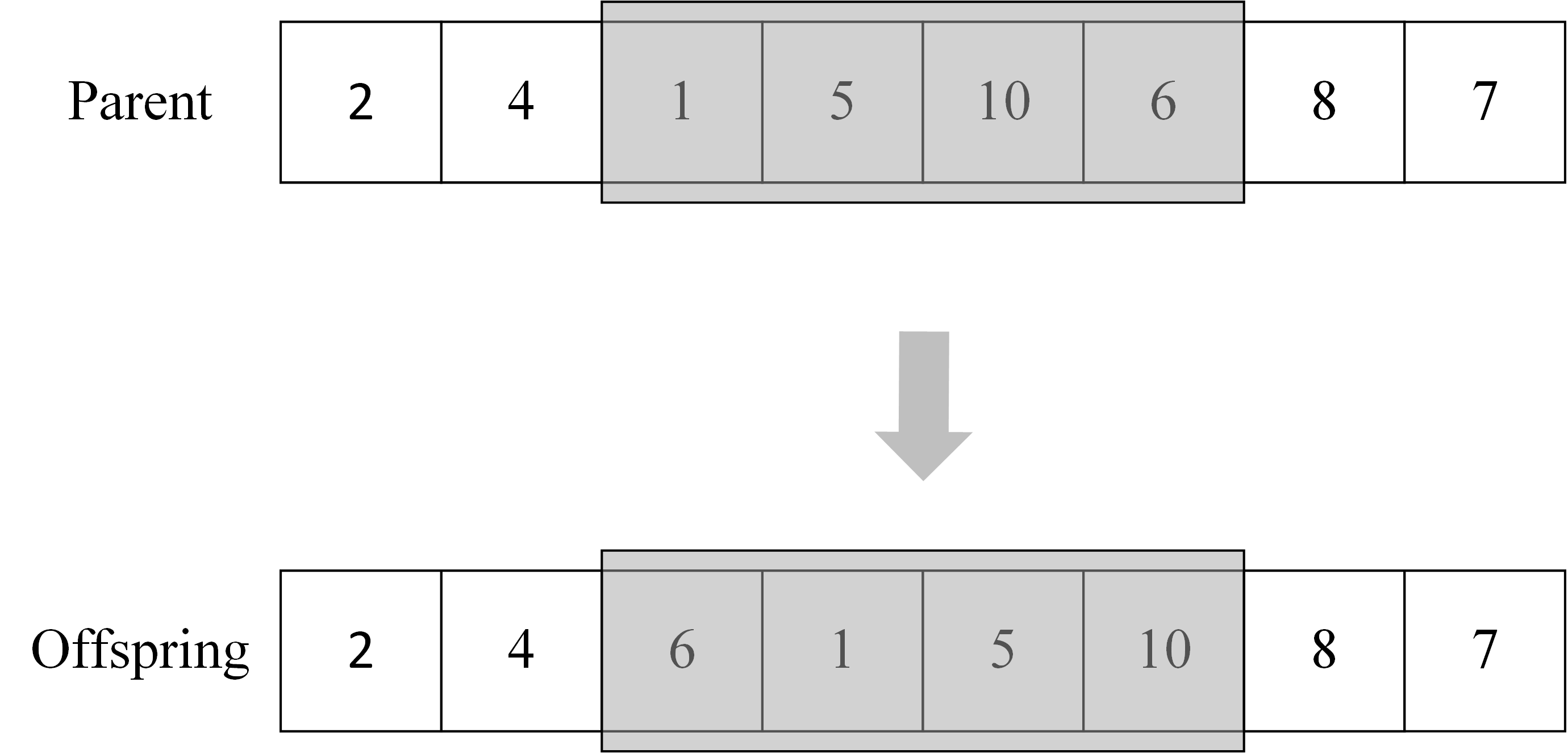}
	\caption{\textcolor[rgb]{0,0,0}{Schematic Diagram of Random Crossover Operator 2.}}
	\label{Schematic Diagram of Random Crossover Operator 2}
\end{figure}

\textbf{Random crossover operator 2}: As shown in Fig.\ref{Schematic Diagram of Random Crossover Operator 2}, randomly select a certain length segment from a task sequence segment, shuffle the original order of the segment, and reinsert it into the corresponding position of the sequence in relative order.

\begin{figure}[htp]
	\centering
	\includegraphics[width=0.4\textwidth]{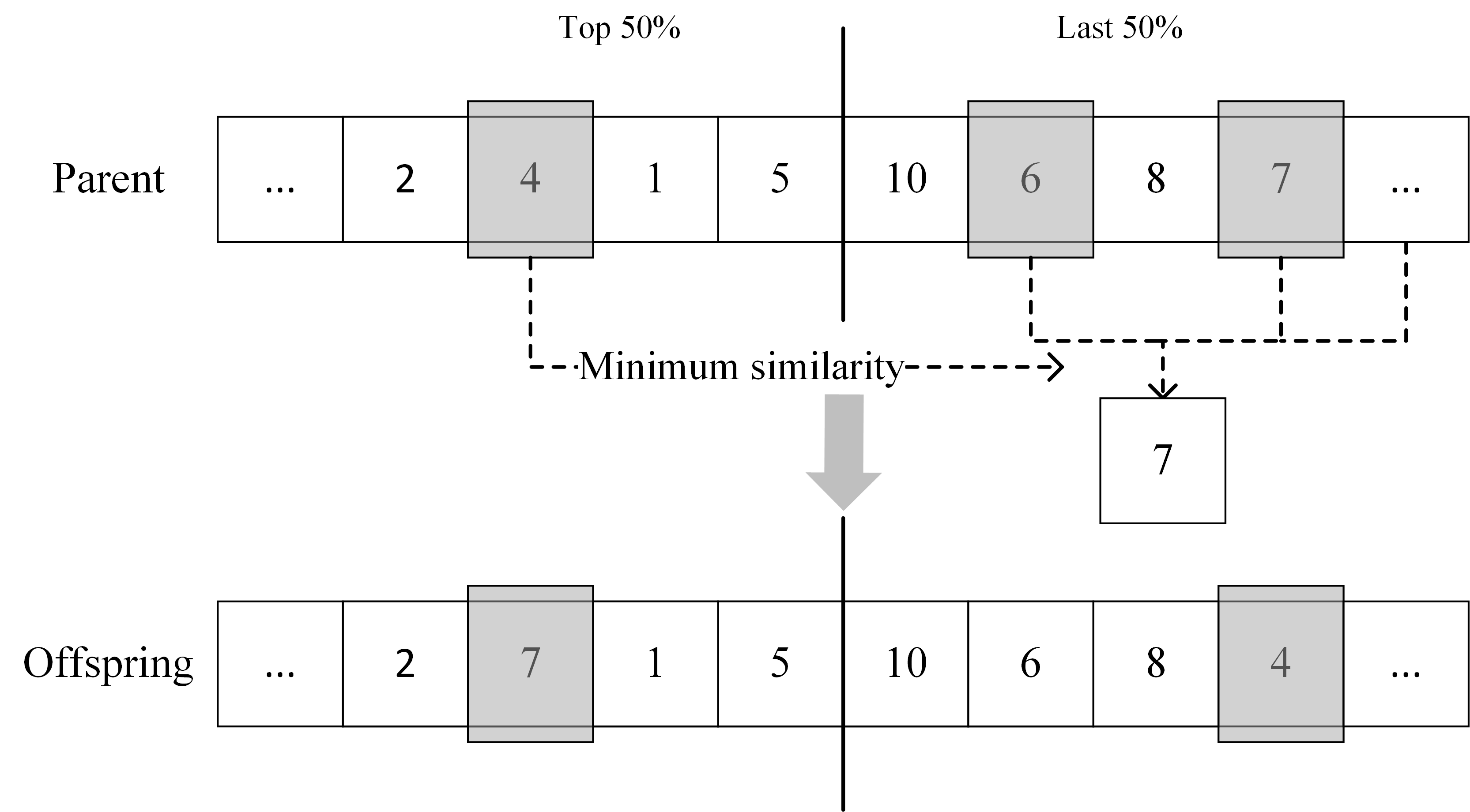}
	\caption{\textcolor[rgb]{0,0,0}{Schematic Diagram of Similarity Crossover Operator 1.}}
	\label{Schematic Diagram of Similarity Crossover Operator 1}
\end{figure}

\textbf{Similarity crossover operator 1}: As shown in Fig.\ref{Schematic Diagram of Similarity Crossover Operator 1}, randomly select a segment from the top 50\% of a task sequence, and then randomly select several different segments from the last 50\% to find the task with the minimum similarity. Swap the positions of these two segments to form a new task sequence.

\begin{figure}[htp]
	\centering
	\includegraphics[width=0.4\textwidth]{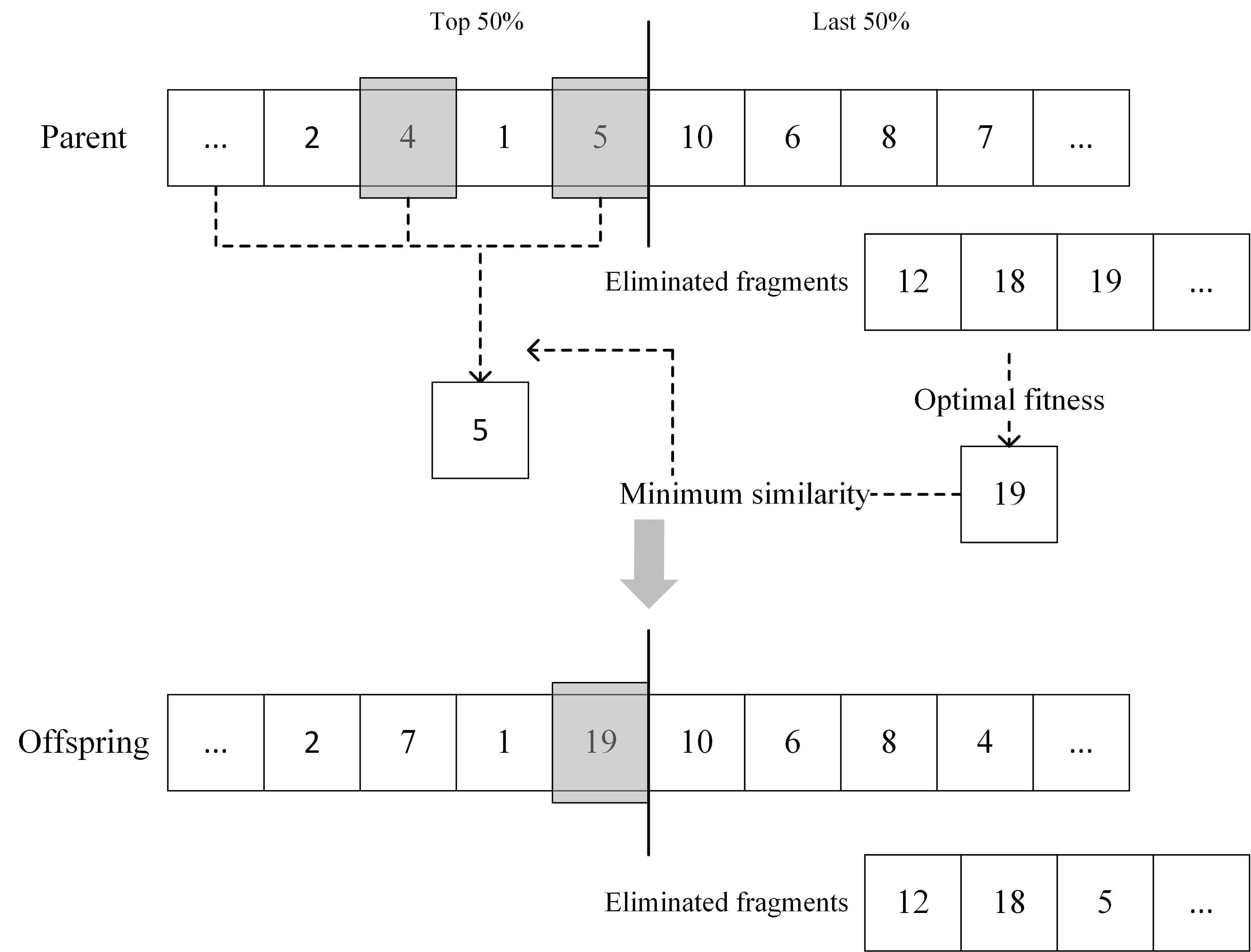}
	\caption{\textcolor[rgb]{0,0,0}{Schematic Diagram of Similarity Crossover Operator 2.}}
	\label{Schematic Diagram of Similarity Crossover Operator 2}
\end{figure}

\textbf{Similarity crossover operator 2}: As shown in Fig.\ref{Schematic Diagram of Similarity Crossover Operator 2}, unscheduled task fragments, which are eliminated fragments, will be obtained in the TSMFS method. Then, find the elimination task $\alpha$ with the highest profit in this segment, and randomly select several different segments from the top 50\% of the task sequence to find the task with the smallest similarity to $\alpha$. Swap the positions of these two segments to form a new task sequence.

The DSGA employs an adaptive selection strategy based on the search performance of each operator and the offspring generated. Prior to iteration, all operators are assigned equal scores, and the weights of crossover operators are determined by calculating their proportion relative to the total score of all operators.

The algorithm employs the RWS method to determine the crossover operator used based on the acquired weight values. Subsequently, each individual within the population will evolve according to the selected operator. Following cross evolution, the algorithm updates the score of the employed crossover operator based on its fitness and determines whether to update the evolved population accordingly. The pseudo-code for score updating is presented in Algorithm \ref{a:score}. After a certain number of score updates, it becomes necessary to update the weights of operators. Consequently, a new operator will be chosen based on these updated weight values. The formula for weight updating is as follows.

\begin{equation}
	w_i=\frac{score_i}{\sum_i^Mscore_i}
\end{equation}

\noindent where $score_i$ represents the score of the $i$-th crossover operator, $w_i$ represents the weight of the $i$-th crossover operator, $M$ represents the number of crossover operators. In this paper, $M=4$. When the number of adaptive optimizations reaches a certain threshold, each operator will update its weights based on the new score.

\begin{algorithm}[htbp]
	\caption{Score update method}
	\label{a:score}
	\small
	\LinesNumbered %要求显示行号
	\KwIn {Global optimal solution $ f_{global} $, Current solution $ f_{local} $, Last iteration solution $ f_{last} $, Score of selected crossover operator $Score_{sel} $, Four scores $ \alpha,\beta,\gamma\,\delta $}
	\If{$f_{local} > f_{global}$}{
		$Score_{sel} \leftarrow Score_{sel} + \alpha$\;
	}
	\ElseIf{$f_{local} > f_{last}$}{
		$Score_{sel} \leftarrow Score_{sel} + \beta$\;
	}
	\ElseIf{$f_{local} > f_{last} \ast $ Metropolis criterion}{
		$Score_{sel} \leftarrow Score_{sel} + \gamma$\;
	}
	\Else{
		$Score_{sel} \leftarrow Score_{sel} + \delta$\;
	}
	
\end{algorithm}

\subsection{Task Scheduling Method with Feed Switching Mode (TSMFS)}
\label{tsmfs}
We propose a task scheduling method, namely the task scheduling method with feed switching mode (TSMFS), which aims to schedule each task optimally in order to transform an individual into an executable task-specific program. The pseudo-code for TSMFS is presented in Algorithm \ref{a:tsmfs}. As depicted in Algorithm \ref{a:tsmfs}, the design of TSMFS is based on addressing the SGNPFM problem. By establishing connections between different time windows, TSMFS can determine the start and end times of tasks, enabling assessment of whether sufficient time exists for their execution. The core concept behind TSMFS involves prioritizing the insertion of time windows (TWs) into tasks to derive their respective start and end times. Subsequently, a set of scheduling rules are employed to evaluate potential task execution plans and perform constraint checks, ultimately determining the final task execution plan. A comprehensive description of TSMFS will be provided in the subsequent paragraph.

\begin{algorithm}[htbp]
	\caption{Task scheduling method with feed switching mode (TSMFS)}
	\label{a:tsmfs}
	\small
	\LinesNumbered %要求显示行号
	\KwIn {Task Set $ T $, Time Window Set $ TW $, Feeding mode Time Window $ FTW $, Population $ P $}
	\KwOut {$ Solution $}%输出
	\ForEach{$indi_o $ in $ P $}
	{
		\ForEach{$task_k $ in the order in $ indi_o $}
		{
			\ForEach{$ tw_{p}^{ijmn} $ in $ TW $}
			{	
				$ aest_k \leftarrow max\left\{est_k,tws_{p}^{ijmn}\right\}$\;
				$ alet_k \leftarrow min\left\{let_k,twe_{p}^{ijmn}\right\}$\;
				\If{$\left(aest_k - alet_k\right) \ge d_k $}
				{	
					$ st_k, et_k \leftarrow $ Arrange task start time and end time at $ aest_k$ and at $\left(st_k+d_k\right)$\;
					Omit $ tw_{p}^{ijmn} $ frow $ TW $\;
					Generate two new time windows $ tw_{p'}^{ijmn} $ and $ tw_{p''}^{ijmn} $ with the attributes $\left[tws_{p}^{ijmn}, st_k\right]$ and $\left[et_k,twe_{p}^{ijmn}\right]$\;
					$ TW \cup \left\{tw_{p''}^{ijmn}, tw_{p''}^{ijmn}\right\} \leftarrow $ Update $ TW $ with new time windows\;
					Try to arrange the next task $task_{k+1}$\; 
				}
				\ElseIf	{$\left(aest_k - alet_k\right) < d_k $}
				{
					$ftws_{q}^{ijmn} \leftarrow $ Find $ftw_{q}^{ijmn}$ associated with $tw_{p}^{ijmn}$\;
					\If{$ftws_{q}^{ijmn} \le alet_k$}
					{
						$ st_k, et_k \leftarrow $ Arrange task start time and end time at $ aest_k$ and at $\left(st_k+d_k\right)$\;
						Omit $ tw_{p}^{ijmn} $ frow $ TW $\;
						Generate two new time windows $ tw_{p'}^{ijmn} $ with the attributes $\left[tws_{p}^{ijmn}, st_k\right]$\;
						Omit $ ftw_{q}^{ijmn} $ frow $ FTW $\;
						Generate two new time windows $ ftw_{q'}^{ijmn} $ with the attributes $\left[et_k, ftwe_{q}^{ijmn}\right]$\;
						$ TW \cup tw_{p'}^{ijmn} \leftarrow $ Update $ TW $ with new time windows\;
						$ FTW \cup ftw_{q'}^{ijmn} \leftarrow $ Update $ FTW $ with new time windows\;
						Try to arrange the next task $task_{k+1}$\; 
					}
				}
				\Else
				{	
					Turn to next time window $tw_{p+1}^{ijmn}$\;
				}
			}
		}
		
	}
\end{algorithm}
Firstly, a task examines a time window to determine its feasibility for sequential execution. Meanwhile, two new variables $aest_k$ and $alet_k$ are introduced prior to scheduling, representing the earliest actual available time and the latest actual end time respectively. These variables are computed according to Eq.\eqref{f:tsmfs1} and Eq.\eqref{f:tsmfs2}. If the duration of a time window exceeds the required task completion time, an attempt is made to schedule the task within that specific window. Subsequently, if the actual availability range surpasses the required time range for task execution, successful scheduling can be achieved. Following successful task allocation, the corresponding time window needs to be inserted based on both its earliest actual start time and latest actual end time; subsequently requiring cropping of this newly inserted interval. Once this cropping process is completed, the time window set will be updated.

\begin{equation}
	\label{f:tsmfs1}
	aest_k =\max \{est_k,tws_p^{ijmn}\}
\end{equation}
\begin{equation}
	\label{f:tsmfs2}
	alet_k =\min \{let_k,twe_p^{ijmn}\}
\end{equation}

If the time required for scheduling a task exceeds the time range that can be scheduled within the current time window, it proves that the task cannot be scheduled within the current time window. After the current time window scheduling is unsuccessful, it is necessary to determine the switching of the feeding mode. Firstly, the regular time window and the feeding time window are correlated through prior knowledge. If there is a time overlap between the two time windows, they will be bound together. When the task fails to schedule within the current time window, it can be determined whether there is an associated feeding time window within the current time window, rather than needing to search again. The purpose of doing so is to reduce the additional resources generated during the search for feeding time windows.

\begin{figure}[htp]
	\centering
	\subfigure[Schematic Diagram of Task Execution Failure]{
		\includegraphics[width=0.4\textwidth]{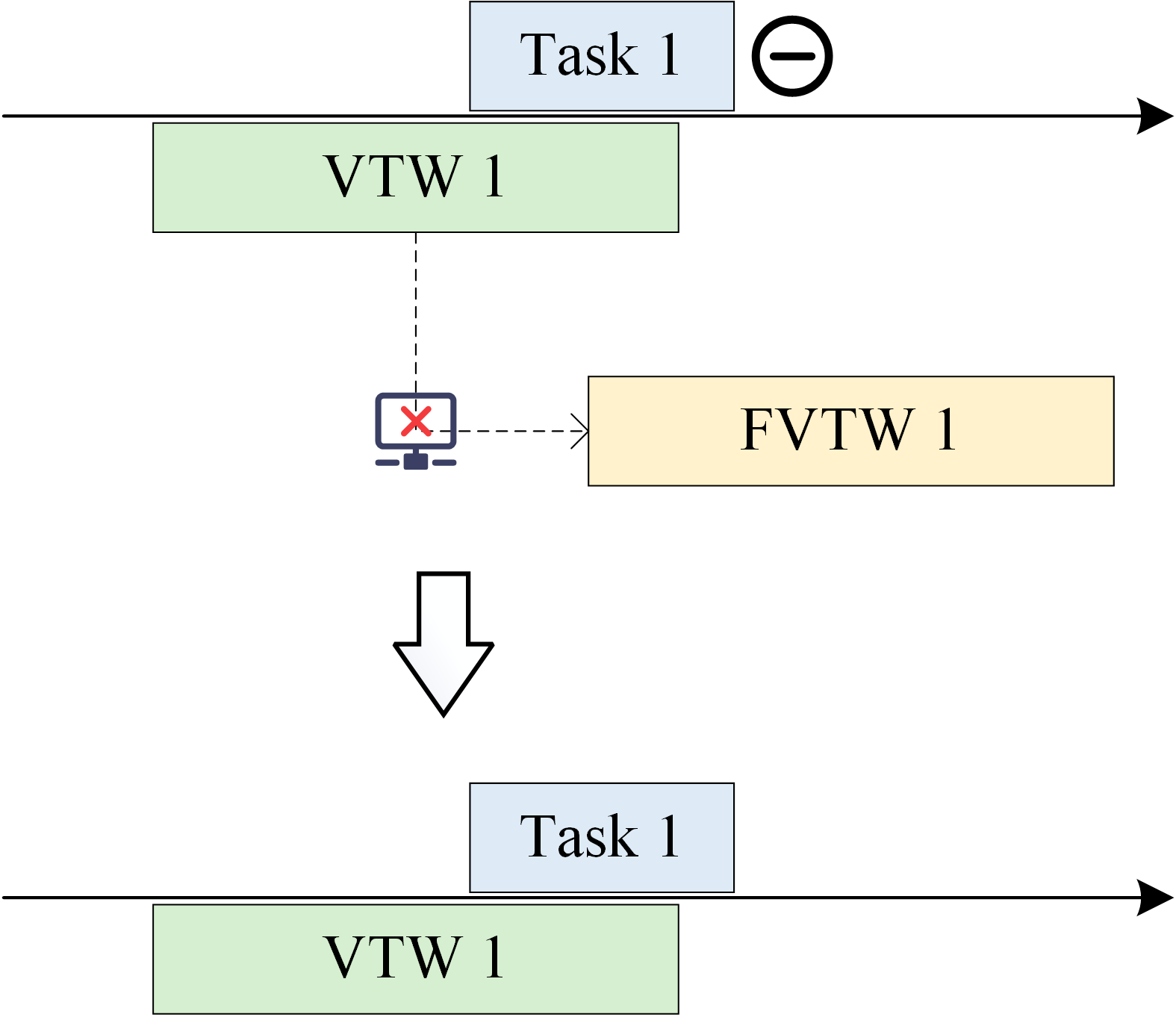}}
	\subfigure[Schematic Diagram of Feed-switching]{
		\includegraphics[width=0.4\textwidth]{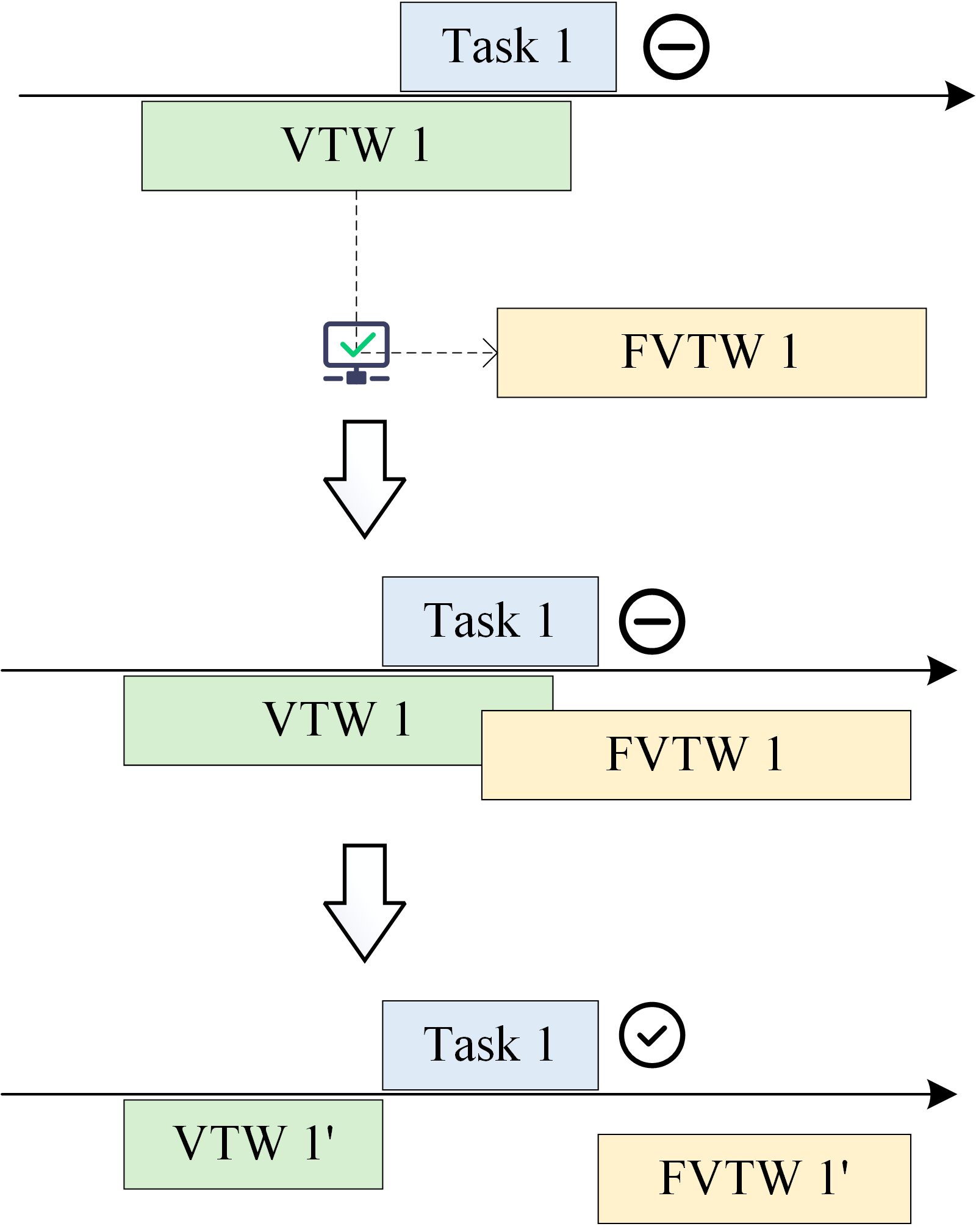}}
		
	\caption{\textcolor[rgb]{0,0,0}{Schematic Diagram of Setup Time Required for Execution between Two Tasks.}}
	\label{Schematic Diagram of Task Scheduling Method}
\end{figure}

If there are associated feeding time windows within the current time window, the task can be scheduled during these two specific periods. The start time of the task will be set to the earliest available time within the regular time window, while the end time will align with the actual end time of the feeding time window. Once successfully scheduled in both windows, they can be adjusted and updated accordingly. Conversely, if no associated feeding time window exists within the current period, scheduling for this particular task is not possible and consideration will be given to scheduling the next task. As shown in Fig.\ref{Schematic Diagram of Task Scheduling Method}, when the duration of the task exceeds the time window and there is no feeding time window to connect, the task execution fails in Fig.13(a); When the task duration exceeds the time window in Fig.13(b), but there is a feeding time window that can be connected, the task is executed successfully and the time window is cropped.

\section{Experimen}
\label{Experimental results and discussions}

\subsection{Experiment Settings}
We carried out the following simulation experiments to verify the DSGA. Various experimental settings are described in the following sections.

\subsubsection{Experiment Environment}
All algorithms in the experiment are run by Matlab2023a under the configuration environment of Core I7-9750 2.6 GHz CPU, 16 GB RAM, and Windows 10 operating system.
\subsubsection{Instance Settings}
As there is no public test set for SGNPFM, we use a random method to generate test cases. The time range of the scenarios is 24 hours, and the number of tasks is between 100-1000. The communications task requirements time follows a normal distribution with a mean of 55 and a standard deviation of 45. The task profits follow a uniform distribution with a mean of 10 and a standard deviation of 9. For convenience, the form "$A-B$" is used to label instances, where $A$ denotes the size of the instance and $B$ denotes the internal number of the instance at that size. According to the number of instances are categorized into three types: small scale, medium scale and large scale, small scale instances are set between 100 $\sim$ 300, medium scale instances are set between 400 $\sim$ 600, and large scale instances are set between 700 $\sim$ 1000.

\subsubsection{Comparative Algorithms}
\textcolor[rgb]{0,0,0}{To test the effectiveness of the proposed algorithm for solving the SGNPFM problem, we use a variety of algorithms that perform well in various combinatorial optimization problems as comparison algorithms. These comparison algorithms are a knowledge-based evolutionary algorithm (KBGA)\cite{song2020knowledge}, an improved adaptive large neighborhood search algorithm (ALNS-I)\cite{he2020time}, artificial bee colony algorithm (ABC)\cite{song2020multi}. The KBGA is an efficient search using domain knowledge driven algorithm. Similar to KBGA, our proposed DSGA uses genetic algorithm as the basic algorithm for solving problems, and also incorporates knowledge from the satellite field into the initialization process. So we consider KBGA as a comparative algorithm. The ALNS-I combines tuba search and adaptive large neighborhood search algorithm to find high-quality solutions through destruction and repair. Our proposed DSGA also considers applying adaptive neighborhood knowledge to the crossover operation of genetic algorithms. Each individual in different generations is influenced by selection operators with different scores and weights. Therefore, we consider ALNS-I as a comparative algorithm. The ABC algorithm selects and designs multiple strategies with different search abilities, and proposes the evolution rate, which is an indicator that fully reflects the adaptability of search strategies. Similar to ABC, our proposed DSGA uses similarity as a search metric and also proposes a series of strategies. ABC, as a highly efficient stochastic optimization algorithm, is very suitable as a comparative algorithm. In order to ensure fairness, we add the task scheduling algorithm TSMFS with feed-switching to the three comparison algorithms.}

\subsubsection{Parameters Settings}
We evaluated the performance of DSGA by setting a series of different parameters, including adaptive threshold and operator score. Ultimately, the following parameters demonstrated strong effectiveness in addressing the SGNPFM problem. Consequently, this paper has opted to employ these parameters for comparative experiment.

Algorithm parameter settings: Set all algorithms to terminate after 5000 times of fitness evaluation. The population size of the algorithm is set to 10, the segment length is set to 2, and the maximum length of the buffer pool is set to 20. The probability of the adaptive crossover strategy is set to 0.8, the threshold is 20, and the score is 50, 30, 10, and 5 based on the performance bonus.

\begin{table*}[]
	\centering
	\caption{Scheduling results of the algorithm on all instances}
	\label{t:result}
	\begin{tabular}{lllllllll}
		\toprule[1pt]
		Instance & \multicolumn{2}{l}{DSGA}               & \multicolumn{2}{l}{KBGA}         & \multicolumn{2}{l}{ABC}          & \multicolumn{2}{l}{ALNS-I}      \rule{0pt}{0.4cm} \\ 
		\cline{2-9} 
		& Best          & Mean(Std.Dev.)         & Best          & Mean(Std.Dev.)   & Best          & Mean(Std.Dev.)   & Best          & Mean(Std.Dev.)   \rule{0pt}{0.4cm} \\ 
		\midrule[0.75pt]
		100-1    & \textbf{725}  & \textbf{725(0)}        & \textbf{725}  & \textbf{725(0)}  & \textbf{725}  & \textbf{725(0)}  & \textbf{725}  & \textbf{725(0)}  \\
		100-2    & \textbf{784}  & \textbf{784(0)}        & \textbf{784}  & \textbf{784(0)}  & \textbf{784}  & \textbf{784(0)}  & \textbf{784}  & \textbf{784(0)}  \\
		100-3    & \textbf{818}  & \textbf{818(0)}        & \textbf{818}  & \textbf{818(0)}  & \textbf{818}  & \textbf{818(0)}  & \textbf{818}  & \textbf{818(0)}  \\
		200-1    & \textbf{1635} & \textbf{1635(0)}       & \textbf{1635} & \textbf{1635(0)} & \textbf{1635} & \textbf{1635(0)} & \textbf{1635} & \textbf{1635(0)} \\
		200-2    & \textbf{1698} & \textbf{1698(0)}       & \textbf{1698} & \textbf{1698(0)} & \textbf{1698} & \textbf{1698(0)} & \textbf{1698} & \textbf{1698(0)} \\
		200-3    & \textbf{1629} & \textbf{1629(0)}       & \textbf{1629} & \textbf{1629(0)} & \textbf{1629} & \textbf{1629(0)} & \textbf{1629} & \textbf{1629(0)} \\
		300-1    & \textbf{2435} & \textbf{2421(8.97)}    & 2397          & 2381(7.16)       & 2397          & 2386.3(5.14)     & 2391          & 2381.9(6.73)     \\
		300-2    & \textbf{2452} & \textbf{2446.1(5.09)}  & 2419          & 2401.9(10.63)    & 2427          & 2413.1(7.50)     & 2418          & 2407.1(5.97)     \\
		300-3    & \textbf{2343} & \textbf{2343(0)}       & \textbf{2343} & 2330.7(7.94)     & 2338          & 2333.6(2.99)     & 2340          & 2334.3(3.43)     \\
		400-1    & \textbf{3245} & \textbf{3245(0)}       & \textbf{3245} & \textbf{3245(0)} & \textbf{3245} & \textbf{3245(0)} & \textbf{3245} & \textbf{3245(0)} \\
		400-2    & \textbf{3280} & \textbf{3280(0)}       & \textbf{3280} & 3277.7(2.54)     & \textbf{3280} & \textbf{3280(0)} & \textbf{3280} & 3278.1(2.73)     \\
		400-3    & \textbf{3247} & \textbf{3247(0)}       & \textbf{3247} & \textbf{3247(0)} & \textbf{3247} & \textbf{3247(0)} & \textbf{3247} & \textbf{3247(0)} \\
		500-1    & \textbf{3972} & \textbf{3972(0)}       & \textbf{3972} & 3968.4(3.24)     & \textbf{3972} & 3971.4(1.35)     & \textbf{3972} & 3967.9(1.60)     \\
		500-2    & \textbf{4073} & \textbf{4065.6(3.47)}  & 4038          & 4026.9(7.69)     & 4044          & 4033.5(5.70)     & 4040          & 4029.1(6.31)     \\
		500-3    & \textbf{4132} & \textbf{4132(0)}       & \textbf{4132} & 4129.1(2.39)     & \textbf{4132} & \textbf{4132(0)} & \textbf{4132} & 4131.9(0.32)     \\
		600-1    & \textbf{4994} & \textbf{4994(0)}       & 4988          & 4978.8(4.66)     & 4994          & 4987(4.06)       & 4994          & 4983.4(6.75)     \\
		600-2    & \textbf{4872} & \textbf{4872(0)}       & 4865          & 4858.6(5.56)     & 4872          & 4871.4(1.35)     & 4872          & 4868.8(4.44)     \\
		600-3    & \textbf{4546} & \textbf{4529.7(5.06)}  & 4525          & 4504.3(10.56)    & 4529          & 4520.7(7.35)     & 4526          & 4517.8(3.79)     \\
		700-1    & \textbf{5694} & \textbf{5683.7(6.41)}  & 5649          & 5633.5(8.67)     & 5665          & 5651.3(6.09)     & 5672          & 5643.9(12.39)    \\
		700-2    & \textbf{5759} & \textbf{5759(0)}       & \textbf{5759} & \textbf{5759(0)} & \textbf{5759} & \textbf{5759(0)} & \textbf{5759} & \textbf{5759(0)} \\
		700-3    & \textbf{5729} & \textbf{5729(0)}       & \textbf{5729} & 5727.9(1.79)     & \textbf{5729} & \textbf{5729(0)} & \textbf{5729} & \textbf{5729(0)} \\
		800-1    & \textbf{6251} & \textbf{6241.9(6.54)}  & 6232          & 6224.9(5.95)     & 6245          & 6236.8(3.97)     & 6240          & 6234.1(2.73)     \\
		800-2    & \textbf{6665} & \textbf{6653.9(9.67)}  & 6621          & 6606.8(12.31)    & 6642          & 6628.6(7.06)     & 6650          & 6616.8(14.81)    \\
		800-3    & \textbf{6483} & \textbf{6477.8(3.29)}  & 6467          & 6453.9(7.05)     & 6477          & 6469.5(4.60)     & 6473          & 6458.1(9.02)     \\
		900-1    & \textbf{7157} & \textbf{7150.2(5.69)}  & 7104          & 7080.4(15.61)    & 7118          & 7100.6(11.75)    & 7121          & 7096.7(14.28)    \\
		900-2    & \textbf{7196} & \textbf{7196(0)}       & 7195          & 7194.3(2.45)     & \textbf{7196} & \textbf{7196(0)} & 7191          & 7186(6.85)       \\
		900-3    & \textbf{7467} & \textbf{7466.7(0.95)}  & 7458          & 7435.6(11.25)    & 7464          & 7450.9(9.53)     & 7456          & 7441.3(9.48)     \\
		1000-1   & \textbf{8019} & \textbf{7999.1(11.95)} & 7914          & 7887.4(19.73)    & 7915          & 7896.8(10.16)    & 7897          & 7881.4(11.43)    \\
		1000-2   & \textbf{7703} & \textbf{7682.7(11.75)} & 7598          & 7578.8(11.95)    & 7644          & 7609.3(15.41)    & 7632          & 7596.3(18.26)    \\
		1000-3   & \textbf{7834} & \textbf{7798.9(15.47)} & 7695          & 7624.2(29.51)    & 7713          & 7669.4(22.43)    & 7660          & 7640.1(16.40)    \\ 
		\bottomrule[1pt]
	\end{tabular}
\end{table*}

\subsubsection{Evaluation Metrics}
The algorithms were executed 10 times in our experiments, and the results of a single run were recorded. Then we counted the maximum value (denoted as Max), the average value (denoted as Ave), and the standard deviation (denoted as Std) of the run results. In addition, the CPU time and convergence speed of the algorithms were also used to evaluate the algorithms.

\subsection{Results}

Experiments verify the performance of the proposed algorithm from three aspects: optimization performance, convergence speed and CPU time.

\subsubsection{Evaluation of Scheduling Performance}

\begin{figure*}[htp]
	\centering
	\subfigure[Result of Instance No.300-1]{
		\includegraphics[width=0.4\textwidth]{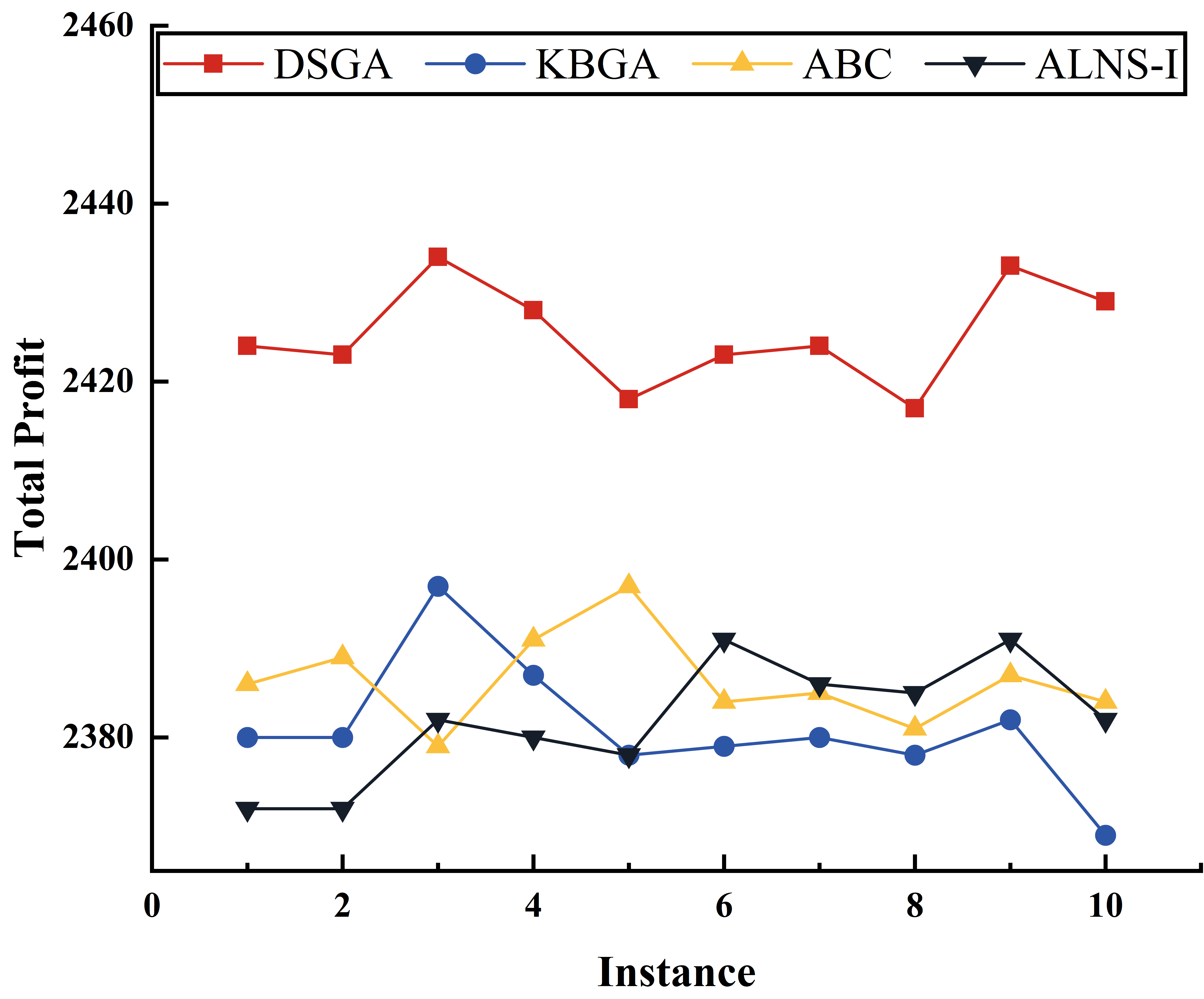}}
	\subfigure[Result of Instance No.500-1]{
		\includegraphics[width=0.4\textwidth]{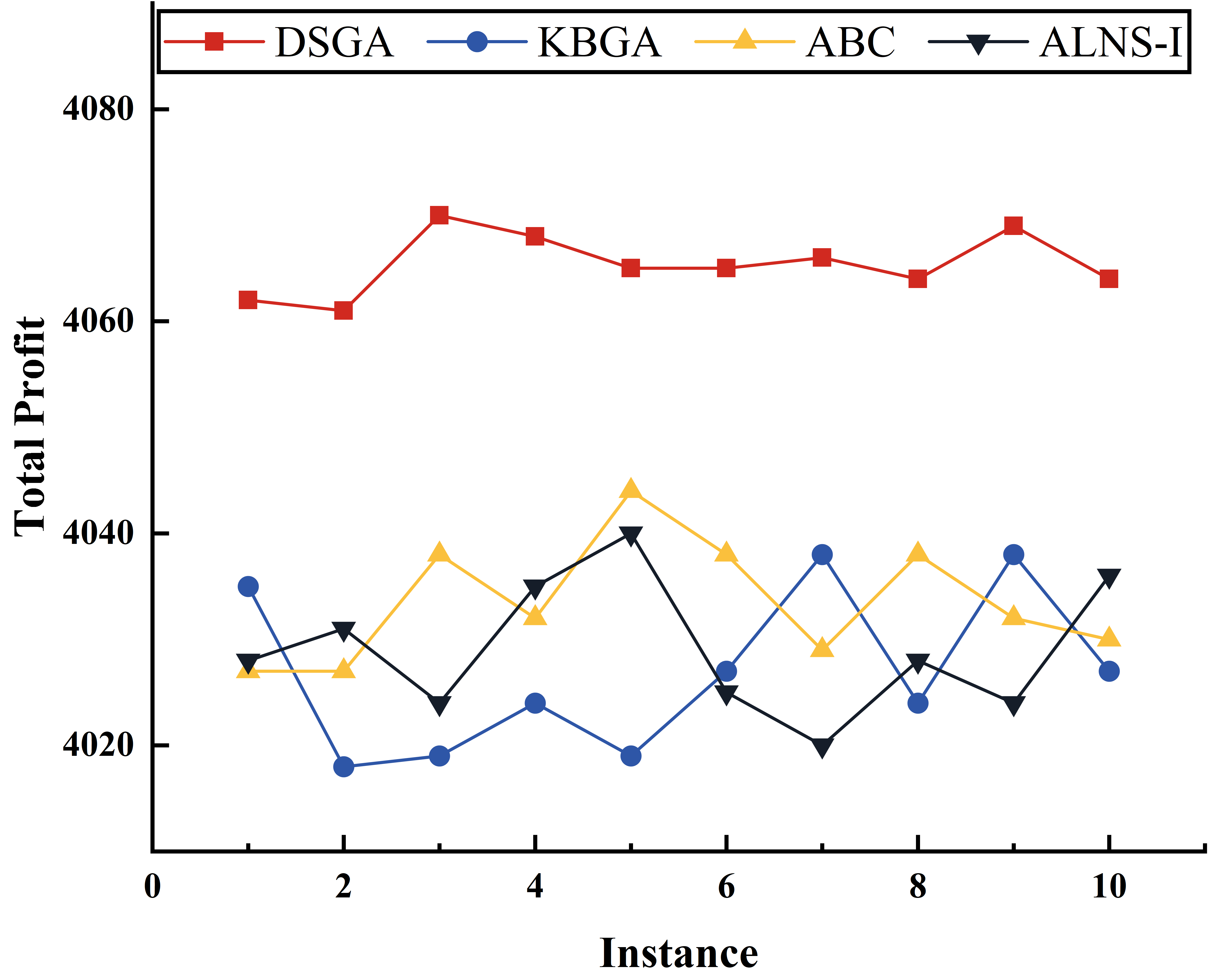}}
	\subfigure[Result of Instance No.800-1]{
		\includegraphics[width=0.4\textwidth]{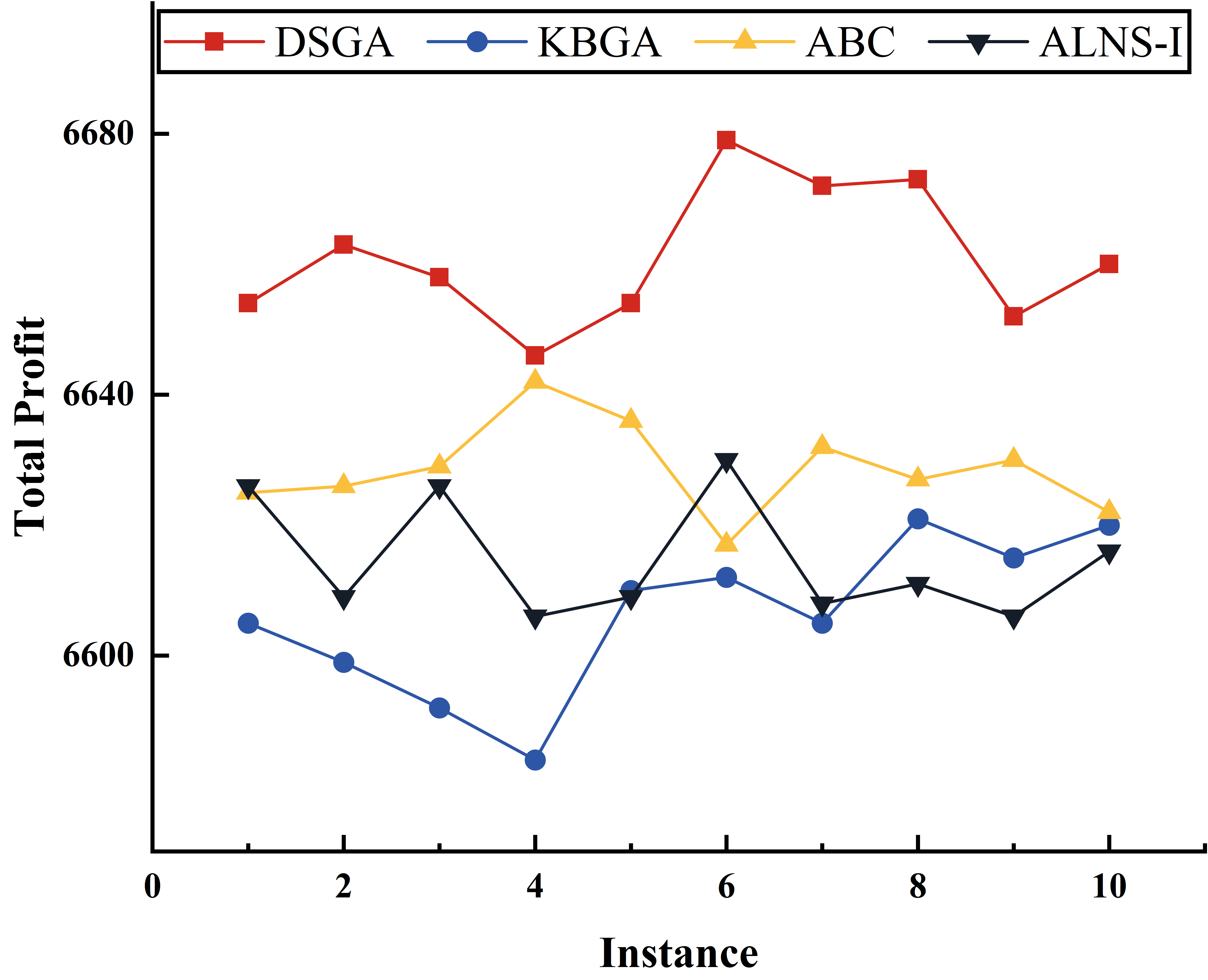}}
	\subfigure[Result of Instance No.1000-1]{
		\includegraphics[width=0.4\textwidth]{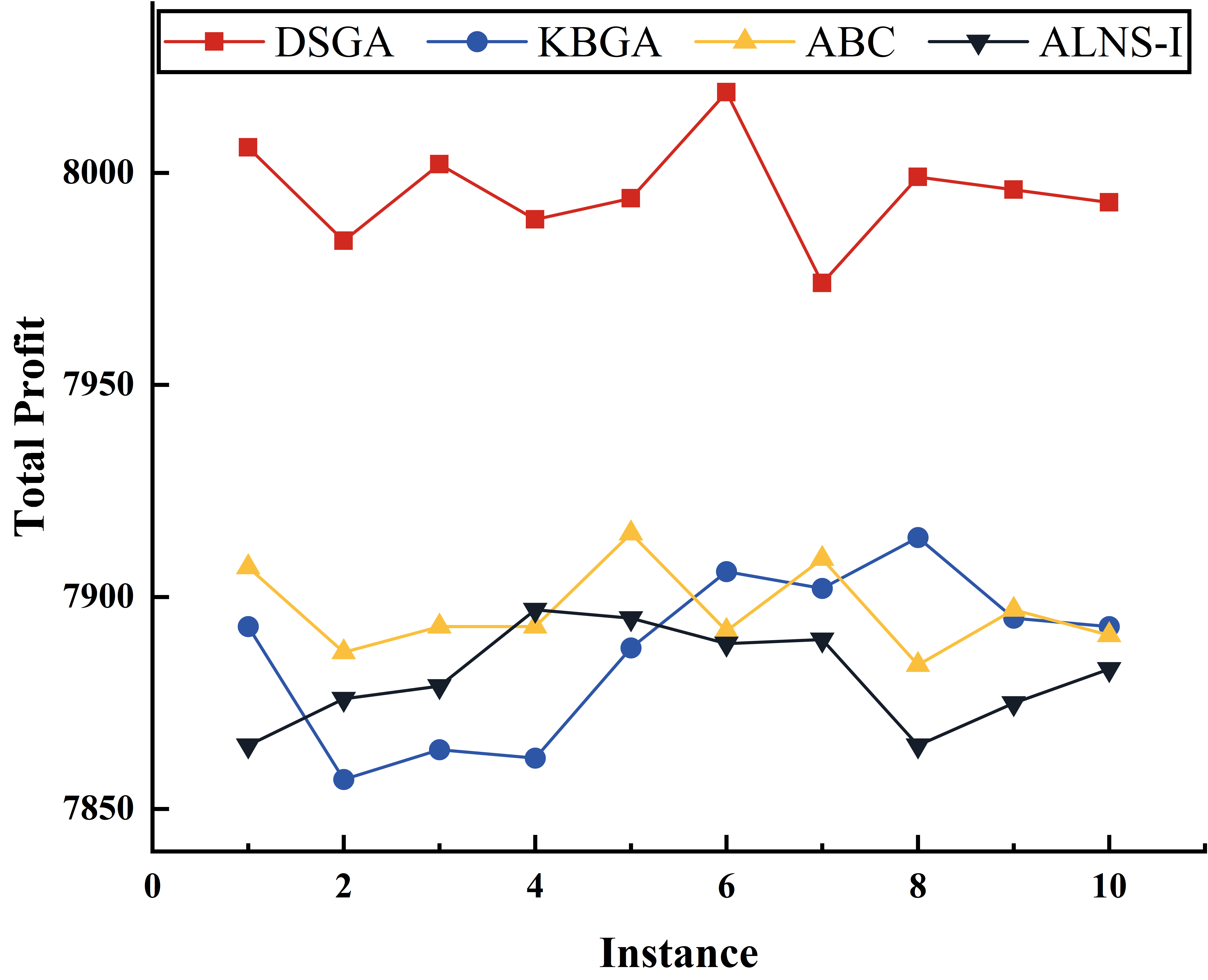}}
	\caption{\textcolor[rgb]{0,0,0}{Result of Different Algorithm Strategies.}}
	\label{Result of Different Algorithm Strategies}
\end{figure*}

Table \ref{t:result} shows the search performance of the algorithms on small-scale to large-scale instances. The optimal solution found is the one with the maximum profit value in the instance. As can be seen from the results, there is no difference in the results of all algorithms in the task with small instances. When the task size increased to 300, the gap between algorithms began to become obvious, and the DSGA found the optimal solution and average solution of most instances. With the increase of task size, the profit gap between the DSGA and other comparison algorithms also increases. Under large-scale instances, our proposed DSGA obtains the maximum detection profit in most cases. There is a clear gap between our proposed algorithm and the other algorithms, which are, in descending order, DSGA, ABC, KBGA, ALNS-I. Fig.\ref{Result of Different Algorithm Strategies} shows more visually the algorithm results of running 10 times at 300-1, 500-1, 800-1 and 1000-1. As can be seen from the figure, the DSGA can obtain higher profits while maintaining good stability.

% \usepackage{multirow}

\begin{figure}[htp]
	\centering
	\includegraphics[width=0.4\textwidth]{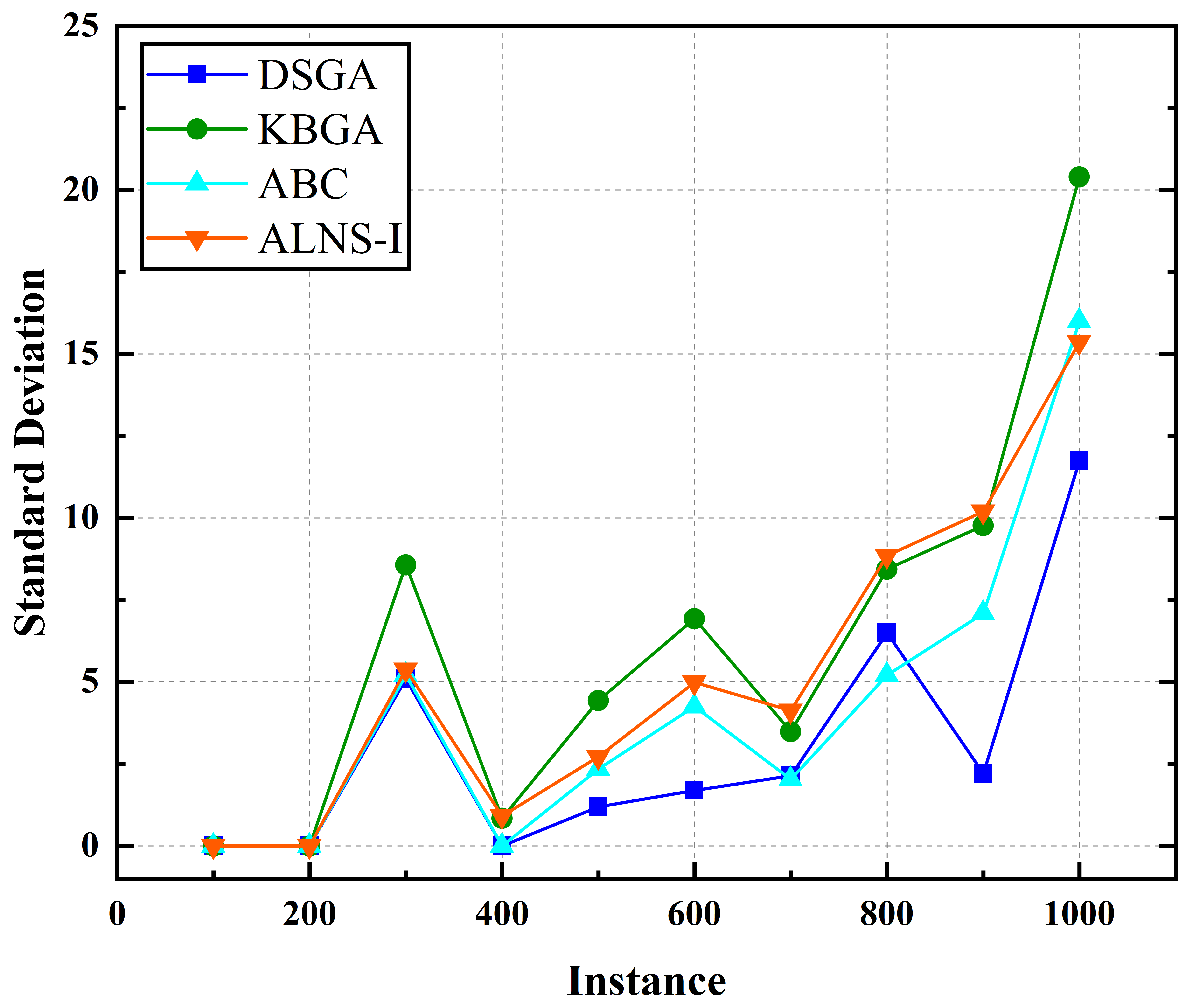}
	\caption{\textcolor[rgb]{0,0,0}{Comparison on The Standard Deviation of Algorithms.}}
	\label{Comparison on The Standard Deviation of Algorithms}
\end{figure}

Fig.\ref{Comparison on The Standard Deviation of Algorithms} shows the average standard deviation of performance for different instance sizes. As can be seen, the DSGA is also excellent in stability. It exhibits not only a consistently low standard deviation in most cases but also demonstrates remarkable stability in search volatility.

\subsubsection{Convergence Analysis}
\begin{figure*}[htp]
	\centering
	\subfigure[Convergence Curves of 500 Task Scale]{
		\includegraphics[width=0.4\textwidth]{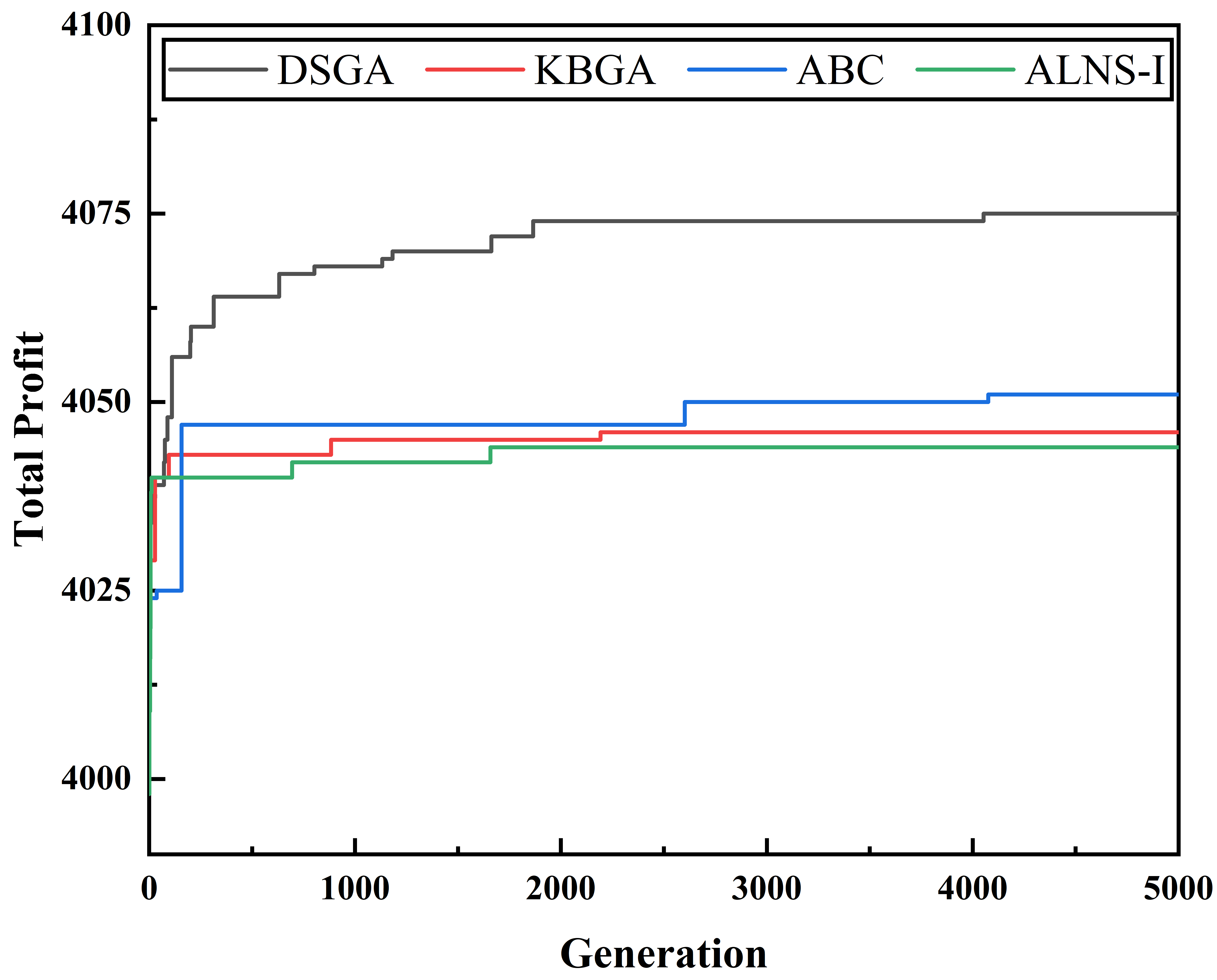}}
	\subfigure[Convergence Curves of 1000 Task Scale]{
		\includegraphics[width=0.4\textwidth]{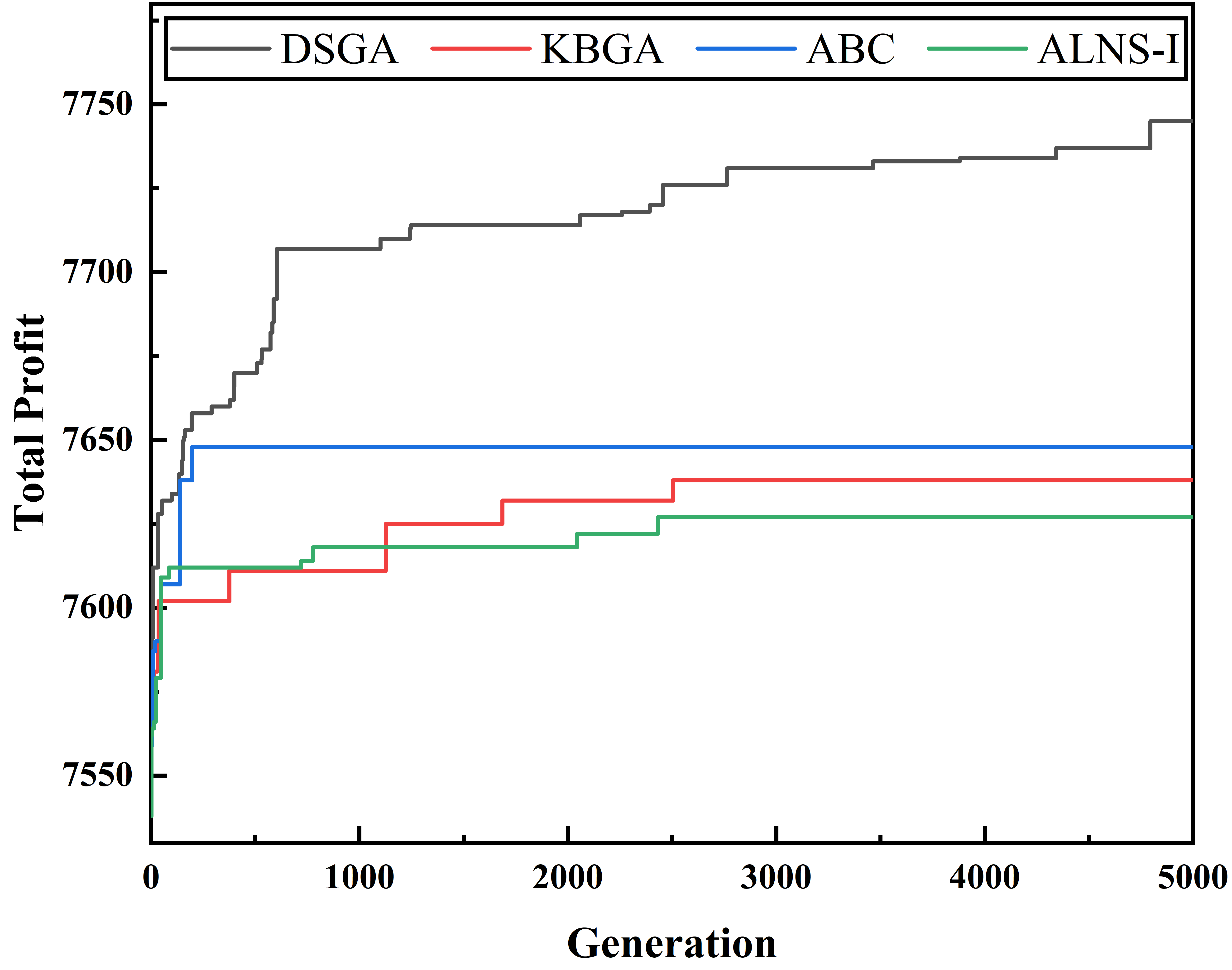}}

	\caption{\textcolor[rgb]{0,0,0}{Convergence Curves of 500 And 1000 Task Scales.}}
	\label{Convergence Curves of 500 And 1000 Task Scales}
\end{figure*}

Fig.\ref{Convergence Curves of 500 And 1000 Task Scales} shows the convergence speed of different algorithms under the task scale of 500-2 and 1000-2 respectively. The results clearly show that in a small-scale instance, when the number of fitness evaluations of all algorithms reaches 3000, the performance of the algorithm is no longer significantly improved. When the task size is large, more search generations are required for the algorithm to converge. Compared with the other three algorithms, the DSGA has a faster convergence speed. The strategy of optimizing the population through the mechanism of similarity ensures that the algorithm does not easily fall into local optimization. Among them, ABC algorithm has faster coverage speed in the initial stage of optimization, but it is easy to fall into local optimization. And KBGA algorithm and ALNS-I coverage is relatively slow. Experimental results show that our proposed algorithm can obtain better optimization performance relative to comparative algorithms while maintaining a better convergence speed.

\subsubsection{CPU Time Analysis}
We then counted the CPU time of the algorithm. Table \ref{t:time} shows the mean CPU time consumption results of the four algorithms on medium and large task instances. From the results, it can be seen that the DSGA has an advantage over the other compared algorithms in terms of search speed. The algorithm exhibits a simple structure and low time complexity in its series of improvements, thereby minimizing the overall impact on time consumption. It is worth noting that the ABC algorithm is characterized by prolonged execution time due to the intricate division of labor and cooperation among the three distinct bees within the population during information search, resulting in extended population iteration. Given that real-world problems often entail complex environments and large task sizes, this paper proposes an algorithm capable of efficient task scheduling.

\begin{table}[htbp]
	\centering
	\caption{Mean CPU time for medium and large instances.}
	\label{t:time}
	\begin{tabular}{lllll}
		\toprule[1pt]
		Instance & DSGA          & KBGA          & ABC   & ALNS-I \\ \midrule[0.75pt]
		400-1    & \textbf{1.38} & \textbf{1.38} & 8.17  & 1.46   \\
		400-2    & \textbf{1.32} & 1.38          & 8.11  & 1.44   \\
		400-3    & \textbf{1.39} & \textbf{1.39} & 7.86  & 1.41   \\
		500-1    & \textbf{1.73} & 1.81          & 10.73 & 1.89   \\
		500-2    & \textbf{1.72} & 1.80          & 10.66 & 1.85   \\
		500-3    & \textbf{1.72} & 1.81          & 10.74 & 1.88   \\
		600-1    & \textbf{2.30} & 2.31          & 13.84 & 2.38   \\
		600-2    & \textbf{2.34} & 2.29          & 13.68 & 2.41   \\
		600-3    & \textbf{2.32} & 2.34          & 14.19 & 2.40   \\
		700-1    & \textbf{3.15} & 3.23          & 19.01 & 3.25   \\
		700-2    & \textbf{2.98} & 3.06          & 18.13 & 3.10   \\
		700-3    & \textbf{3.08} & 3.16          & 18.77 & 3.19   \\
		800-1    & \textbf{3.69} & 3.76          & 22.37 & 3.78   \\
		800-2    & \textbf{3.61} & 3.71          & 21.52 & 3.70   \\
		800-3    & \textbf{3.55} & 3.67          & 21.63 & 3.69   \\
		900-1    & \textbf{4.34} & 4.40          & 25.71 & 4.39   \\
		900-2    & \textbf{4.18} & 4.21          & 24.76 & 4.19   \\
		900-3    & \textbf{4.29} & 4.38          & 25.81 & 4.31   \\
		1000-1   & \textbf{5.01} & 5.15          & 30.18 & 5.11   \\
		1000-2   & \textbf{4.88} & 4.99          & 29.17 & 4.97   \\
		1000-3   & \textbf{5.14} & 5.29          & 30.96 & 5.25   \\ 
		\bottomrule[1pt] 
	\end{tabular}
\end{table}

\subsubsection{Strategy Comparison Analysis}
\begin{figure*}[htp]
	\centering
	\subfigure[Strategy Comparison of 600 Task Scale]{
		\includegraphics[width=0.4\textwidth]{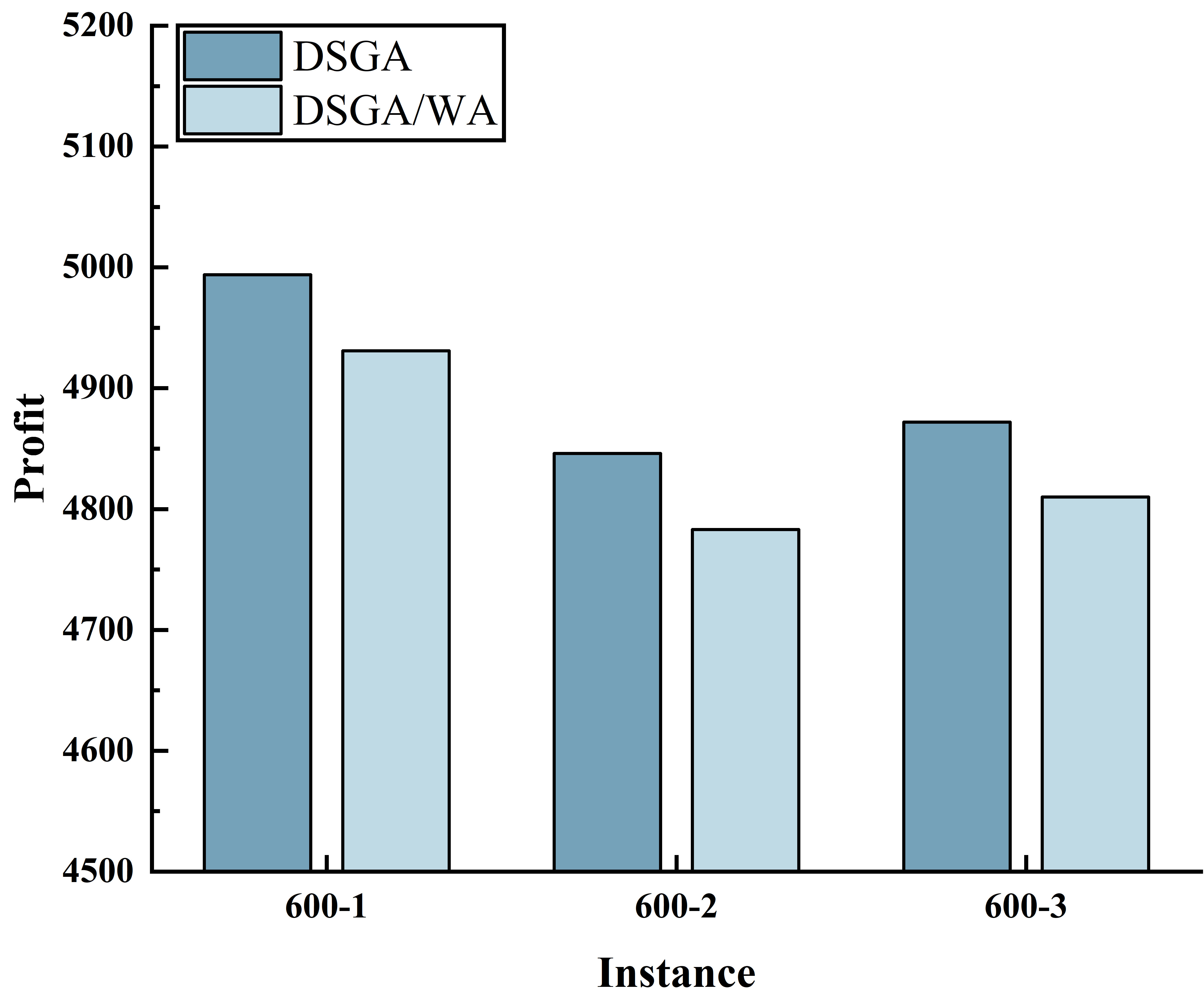}}
	\subfigure[Strategy Comparison of 1000 Task Scale]{
		\includegraphics[width=0.4\textwidth]{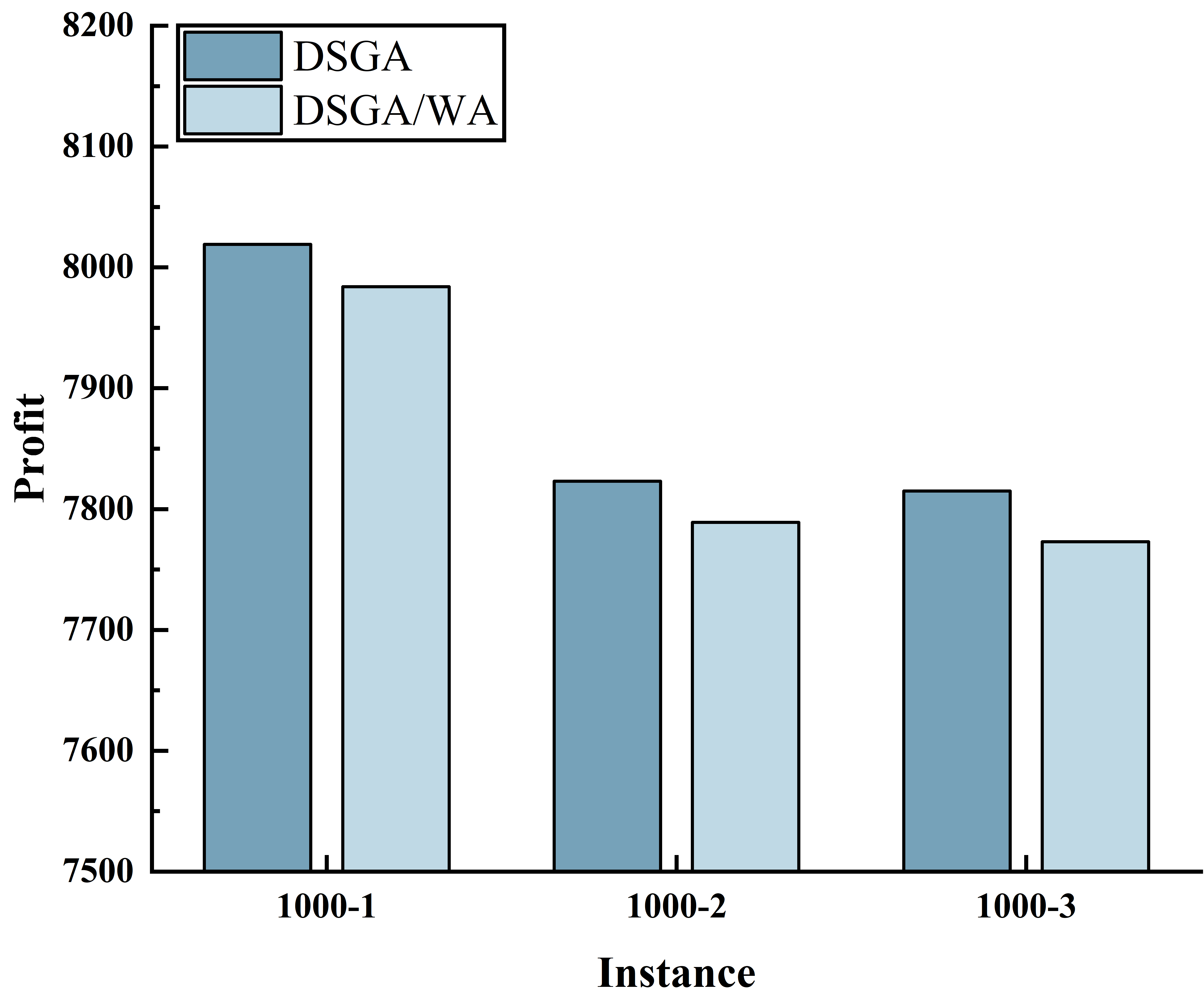}}
	
	\caption{\textcolor[rgb]{0,0,0}{Strategy Comparison Results of Different Scale Instances.}}
	\label{Strategy Comparison Results of Different Scale Instances}
\end{figure*}

We also test whether an adaptive crossover strategy based on similarity improvement can work in the DSGA. Two instances with task sizes of 600 and 1000 are selected to compare DSGA with DSGA without the adaptive crossover strategy (denoted as DSGA/ WA). The average result of profit is shown in Fig.\ref{Strategy Comparison Results of Different Scale Instances}. It can be seen that the DSGA that includes adaptive crossover strategies can always obtain higher profits. Therefore, this shows that the adaptive crossover strategies play a crucial role in the iterative search process of DSGA.

\subsubsection{Discussion}
From the above experimental results, it can be seen that the algorithm can effectively solve the scheduling problem in large-scale task scenarios. Heuristic rules, similarity mechanism, and adaptive cross optimization search effectively improve the search efficiency of genetic algorithm, and the scheduling results are more ideal. Our proposed distance similarity-based genetic optimization algorithm obtains good optimization performance under solving real engineering problems including the studied SGNPFM, which provides a new idea for solving optimization problems in the future.	

In this algorithm, the similarity mechanism is designed to evaluate the impact of the task's attributes on the search, and also to bring a more efficient optimization strategy for the population iteration. Distance similarity can significantly shorten the time of fitness value when the problem is complex, and it can still find high-quality solutions, and finally make more high-quality solutions concentrated. The adaptive crossover operator combined with similarity design also has good performance for the algorithm. Different operators are integrated into the adaptive crossover strategy, and each operator has the possibility to be selected, which increases the diversity of search strategies, and the operator with good performance can explore the solution space more efficiently. Overall, the proposed DSGA can be extended and applied to solve other combinatorial optimization problems in various fields. The distance similarity evaluation mechanism can be further designed. In the future, we can try to use different decision-making methods to determine the specific use of the similarity mechanism to make the results more accurate. Different selection mechanisms can help the algorithm find better strategies.

\section{Conclusion}
\label{Conclusion}
Satellite ground network planning considering feeding mode is a problem of how to improve the efficiency of data transmission during the establishment of the link between the satellite and the ground station, which has rarely been studied by other scholars. In this study, we investigate the SGNPFM problem and develop a mixed integer programming model that incorporates various factors such as feed-switching and antenna link building time. Subsequently, we propose a genetic optimization algorithm based on distance similarity to address the SGNPFM problem. The algorithm combines distance similarity with genetic evolutionary algorithms, enabling it to adapt effectively in diverse scenarios. During the search process, the algorithm efficiently filters out more similar and superior individuals through similarity assessment to enhance search performance. Additionally, an adaptive crossover strategy based on improving similarity is employed for more efficient local search within the algorithm. As evidenced by simulation experiments, the proposed algorithm has demonstrated a nearly 5\% improvement in profit maximization compared to the benchmark algorithm.

In future research, we will extensively investigate the SGNPFM problem in specific scenarios, such as equipment failure of satellites or ground stations, multiple antenna operation modes for satellite feeding, the necessity to establish communication links in a specific period of time, and other complex situations. These practical complexities impose higher demands on algorithm design. The utilization of artificial intelligence algorithms can facilitate more innovative problem-solving approaches. Machine learning methods are well-suited for distributing and evaluating new problems. Reinforcement learning can offer more sophisticated and efficient decision support for algorithms. Furthermore, additional strategies for algorithm improvement can be employed to expedite search processes, including small-scale population and heuristic initialization methods.

\section*{Acknowledgements}

This work is supported by the National Natural Science Foundation of China (723B2002), the Guangxi Intelligent Digital Services Research Center of Engineering Technology, Key Laboratory of Parallel, Distributed and Intelligent Computing (Guangxi University), Education Department of Guangxi Zhuang Autonomous Region Open Subject (IDSOP2305), the National Natural Science Foundation of China (No. 62072124), the Natural Science Foundation of Guangxi (No. 2023JJG170006).

\section*{Declaration of Competing Interest}
The authors declare that they have no known competing financial interests or personal relationships that could have appeared to influence the work reported in this paper.


\bibliographystyle{unsrt}

\bibliography{dsgabib.bib}


%\tableofcontents
	
\end{document}